%% file: main.tex
\documentclass[sigconf, 10pt]{acmart} 

\usepackage{xcolor}
\usepackage{xurl}
\usepackage{booktabs}
\usepackage[ruled]{algorithm2e}
\usepackage{algpseudocode}
\usepackage{enumerate}
\usepackage{enumitem}
\usepackage[english]{babel}
\usepackage{blindtext}
\usepackage[caption=false,font=footnotesize,subrefformat=parens]{subfig}
\usepackage{amsmath}

\usepackage{amssymb}
\usepackage{xspace}
\usepackage{multirow}
\usepackage{threeparttable}
\usepackage{float}
\usepackage{flafter}
\usepackage{comment}
\usepackage[skip=12pt]{caption}
\usepackage{graphicx}
\usepackage{pifont}
\usepackage{tabularx}
\usepackage{bbm}

\def\fig{Fig.\xspace}
\def\eqn{Eq.\xspace}

\def\tab{Tab.\xspace}

\def\ie{{\textit{i.e.}\xspace}} 
\def\eg{{\textit{e.g.}\xspace}}
 
\def\etal{{\textit{et al.}\xspace}}
\def\etc{{\textit{etc}\xspace}} 

\newcommand{\head}[1]{{\noindent \textbf{#1:}}}

\graphicspath{{./figures/}}
\graphicspath{{./pdf/}}
\DeclareGraphicsExtensions{.pdf,.jpeg,.png}

\acmSubmissionID{\#2174, 12+7 pages}

\def\sysname{\textsc{TAP3D}\xspace}

\makeatletter
\def\@ACM@checkaffil{}
\makeatother

\begin{document}

\acmYear{2026}\copyrightyear{2026}
\setcopyright{cc}
\setcctype[4.0]{by}
\acmConference[MobiCom '26]{The 32nd Annual International Conference on Mobile Computing and Networking}{October 26--30, 2026}{Austin, TX, USA}
\acmBooktitle{The 32nd Annual International Conference on Mobile Computing and Networking (MobiCom '26), October 26--30, 2026, Austin, TX, USA}
\acmDOI{10.1145/3795866.3844491}
\acmISBN{979-8-4007-2505-0/26/10}

\title[\sysname]{
\sysname: Thermal-Assisted 3D Human Point Clouds
}

\begin{anonsuppress}
	\author{Xie Zhang}
	\affiliation{%
		\institution{The University of Hong Kong}
	}
	\email{zhangxie@connect.hku.hk}

    \author{Chengxiao Li}
	\affiliation{%
		\institution{The University of Hong Kong}
	}
	\email{chengxiaoli@connect.hku.hk}

    \author{Xuan Liu}
	\affiliation{%
		\institution{The University of Hong Kong}
	}
	\email{etmphile@connect.hku.hk}

	\author{Chenshu Wu}
    \authornote{Corresponding author}
	\affiliation{%
		\institution{The University of Hong Kong}
	}
	\email{chenshu@cs.hku.hk}
\end{anonsuppress}

\renewcommand{\shortauthors}{Zhang \etal}

\begin{abstract}
Human body point clouds are a versatile representation for AI-enabled human sensing. 
However, existing methods using LiDAR, radar, and depth cameras suffer from inherent drawbacks in high cost, sparse reconstruction, and privacy concerns, \etc. 
In this paper, we exploit low-cost thermal arrays and present \sysname, the first system to reconstruct 3D human point clouds from body heat signatures, offering significant advantages in cost, density, human sensitivity, and privacy. 
To overcome major challenges in depth estimation, thermal interference, and multi-person separation, we propose a novel physics-informed design, which integrates a \textit{forward thermal physics model} with two distinct modules: \textit{multi-primitive estimation} for self-supervised joint recovery of depth and other thermal properties, and \textit{geometric perspective fusion} for suppressing interference and disentangling multiple people.
We implement \sysname using a single commodity thermal array sensor and build a large-scale dataset (160K samples, 8 environments, 11 users) for evaluation. 
\sysname achieves remarkable accuracy for dense point cloud generation, enabling downstream tasks like fall detection (91.46\%), indoor tracking (21.86 cm MAE), and human mesh recovery (4.87 cm error). 
By transforming body heat into point clouds for the first time, \sysname pioneers a new paradigm for privacy-first, fully passive human sensing for many applications. 
\sysname is open-sourced at \href{https://github.com/aiot-lab/TAP3D}{\color{blue}\texttt{https://github.com/aiot-lab/TAP3D}}.
\end{abstract}



\maketitle

\input{body_new}

\bibliographystyle{ACM-Reference-Format}
\bibliography{refs}

\newpage
\input{appendix}

\end{document}

%% file: body_new.tex

\begin{figure}[h]
  \includegraphics[width=0.48\textwidth]{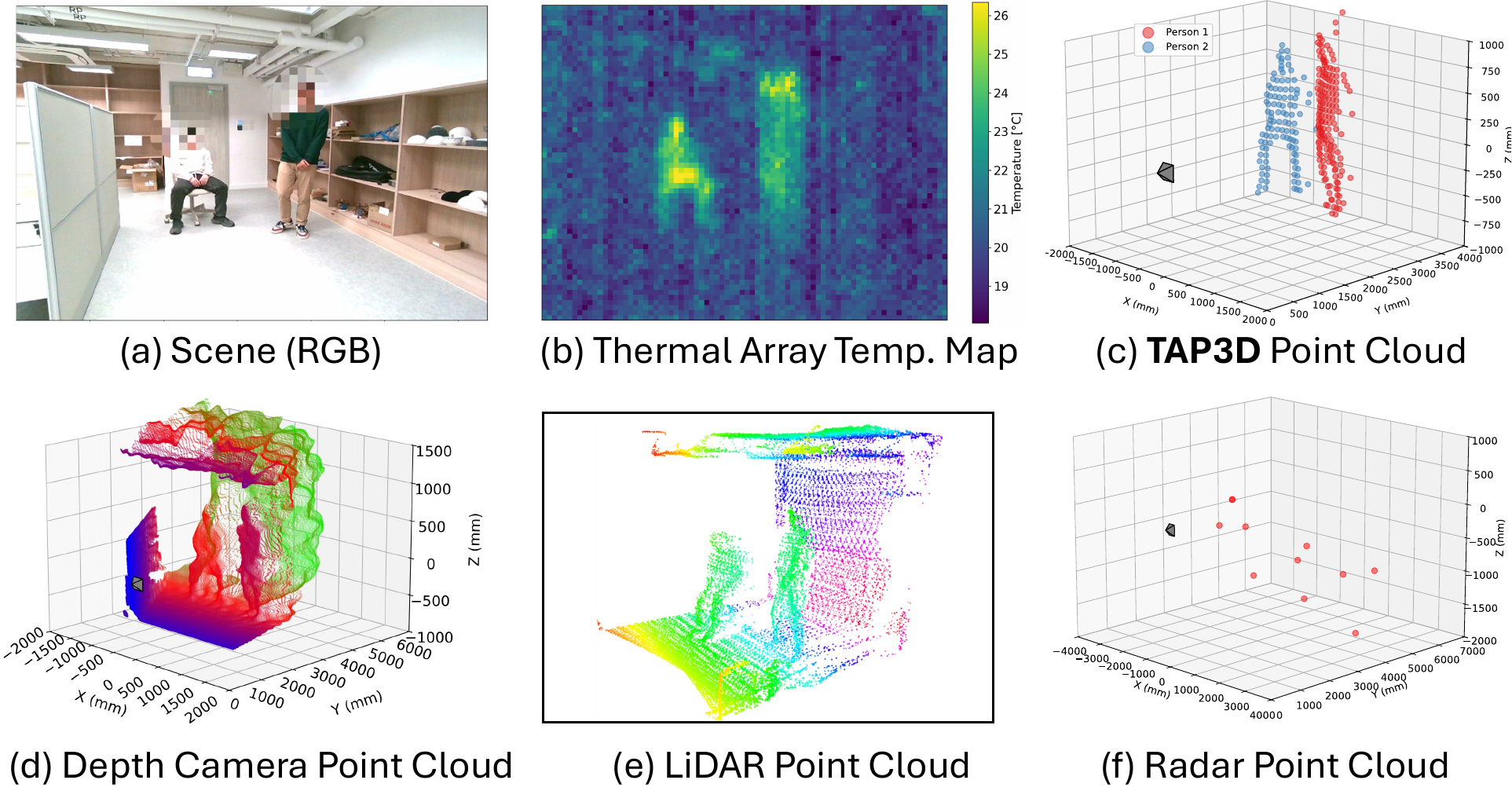}
  \vspace{-0.8\baselineskip}
\caption{Comparison of human point clouds generation across sensing modalities. 
  \rm{\sysname transforms human body (a) heat signatures, captured as a temperature (Temp.) map (b) from a low-cost thermal array sensor, into per-user point clouds (c). 
  For comparison, (d) shows a point cloud from a RealSense stereo vision depth camera; (e) shows a point cloud from a Livox Mid-360 LiDAR; 
  and (f) shows a radar point cloud using TI IWR1843.}}
  \label{fig:modality_overview}
\end{figure}

\section{Introduction}
\label{sec:intro}
Human sensing is fundamental to human-centric applications in healthcare, interactive gaming, augmented and virtual reality, and human–robot interaction. 
\begin{table}[t]
  \centering
  \centering
  \footnotesize
  \caption{Comparison of sensing modalities for human body point clouds generation. \rm{C/R: cost-resolution; Cross-Int.: cross-device interference.}}
  \label{tab:modality_comparison}
  \vspace{-0.1\baselineskip}
  \begin{tabular}{lccccc}
  \hline
  \textbf{Modality}      & \textbf{\begin{tabular}[c]{@{}c@{}}RGB\\ Cam\end{tabular}}      & \textbf{\begin{tabular}[c]{@{}c@{}}Depth\\ Cam\end{tabular}} & \textbf{LiDAR} 
  & \textbf{Radar}
  & \textbf{\begin{tabular}[c]{@{}c@{}}Thermal\\ Array\end{tabular}} \\ \hline
  C/R Balance    & \ding{51}  & \ding{55}                                                               & \ding{55}              & \ding{55}                                                               & \ding{51}                                                                 \\
  Privacy Preserving        & \ding{55}   & \ding{55}                                                               & \ding{51}               & \ding{51}                                                                & \ding{51}                                                                 \\
  Human Sensitive          & \ding{51}    & \ding{51}                                                                & \ding{55}              & \ding{55}                                                               & \ding{51}                                                                 \\
  Smoke/Fog Robust          & \ding{55}    & \ding{55}                                                               & \ding{55}              & \ding{51}                                                                & \ding{51}                                                                 \\
  No Emission Risk         & \ding{51}       & \ding{51}                                                                & \ding{55}              & \ding{55}                                                               & \ding{51}                                                                 \\
  Light Immunity       & \ding{55}  & \ding{55}                                                               & \ding{55}              & \ding{51}                                                                & \ding{51}                                                                 \\
  No Cross-Int. & \ding{51} & \ding{51}                                                                & \ding{55}              & \ding{55}                                                               & \ding{51}                                                                 \\ \hline
  \end{tabular}
\end{table}
Different sensing modalities produce distinct signal representations, for example, range profiles and Doppler spectrum from radar, reflectance maps from LiDAR, and depth maps from stereo vision or ToF-based depth cameras. 
Among them, human body point clouds stand out as a unified and versatile \textit{intermediate representation}, capturing detailed pose and shape information and supporting diverse downstream tasks, such as activity recognition \cite{singhRadHARHumanActivity2019, ulalamPALMARAdaptiveMultiinhabitant2021}, fall detection \cite{bouazizi20212, zhangLTFallDesignImplementation2023, mastorakisFallDetectionSystem2014}, pose estimation \cite{zhangNeighborhoodenhanced3DHuman2024, liLiDARCapLongrangeMarkerless2022a, wangHumanParsingJoint2023}, and mesh reconstruction \cite{fanLiDARHMR3DHuman2023, renLiveHPSLiDARBasedSceneLevel2024, xueM4eshMmWaveBased3D2023a, xueMmMesh3DRealtime2021}, among others.
Point clouds also serve as a \textit{bridge representation} for multi-sensor and cross-modality fusion \cite{abdelazeemMultisensorPointCloud2021, heRegistrationFusionMethod2024, liuBEVFusionMultiTaskMultiSensor2024}, for instance, aligning LiDAR and radar point clouds, a key enabler for embodied AI \cite{ruanSurveyMultisensorFusion2025}.

Different sensing modalities have been explored for point cloud generation, such as depth cameras \cite{sultaniKinect3DPoint2015, xieMozartMobileToF2023, sunThreeDimensionalPointCloud2019}, LiDAR \cite{guoLiDARNetRealscanned3D2024, liDeepLearningLiDAR2021}, and millimeter-wave (mmWave) radar \cite{qian3DPointCloud2020a, prabhakaraHighResolutionPoint2023}. 
However, each modality exhibits inherent limitations such as privacy concerns, low sensitivity to human targets, high costs, and sparse point clouds.
Depth cameras can produce dense and accurate point clouds (see \fig\ref{fig:modality_overview}(d)), yet are sensitive to illumination changes and raise potential privacy concerns \cite{xieMozartMobileToF2023}.
LiDAR provides high-resolution point clouds (see \fig\ref{fig:modality_overview}(e)); however, commercial 3D scanning LiDAR systems remain costly \cite{renLiveHPSLiDARBasedSceneLevel2024, daiHSC4DHumanCentered4D2022}, require additional human detection steps \cite{fanLiDARHMR3DHuman2023}, and pose potential safety risks under prolonged laser exposure in continuous monitoring scenarios \cite{hadlerRandomTestingReveals2013, zuclichOcularEffectsPenetrating1995}. 
Similarly, mmWave radars are privacy-friendly and robust to lighting, but they suffer from cross-device interference in multi-sensor settings \cite{kuiInterferenceAnalysisMmWave2021} and typically yield very sparse point clouds due to the insufficient resolution\footnote{Unless otherwise specified, ``resolution'' refers to \textit{imaging/angle resolution} (spatial sampling density per FoV), rather than \textit{range/depth resolution} (minimum separable distance along the sensing axis).} \cite{luSeeSmokeRobust2020} (see \fig\ref{fig:modality_overview}(f)). 
Enhancing their spatial fidelity often requires more expensive hardware or complicated post-processing \cite{prabhakaraHighResolutionPoint2023, zhangMmFERMillimetrewaveRadar2023a, liHighresolutionHandheldMillimeterwave2024}.

\begin{figure}[t]
  \centering
 \includegraphics[width=0.46\textwidth]{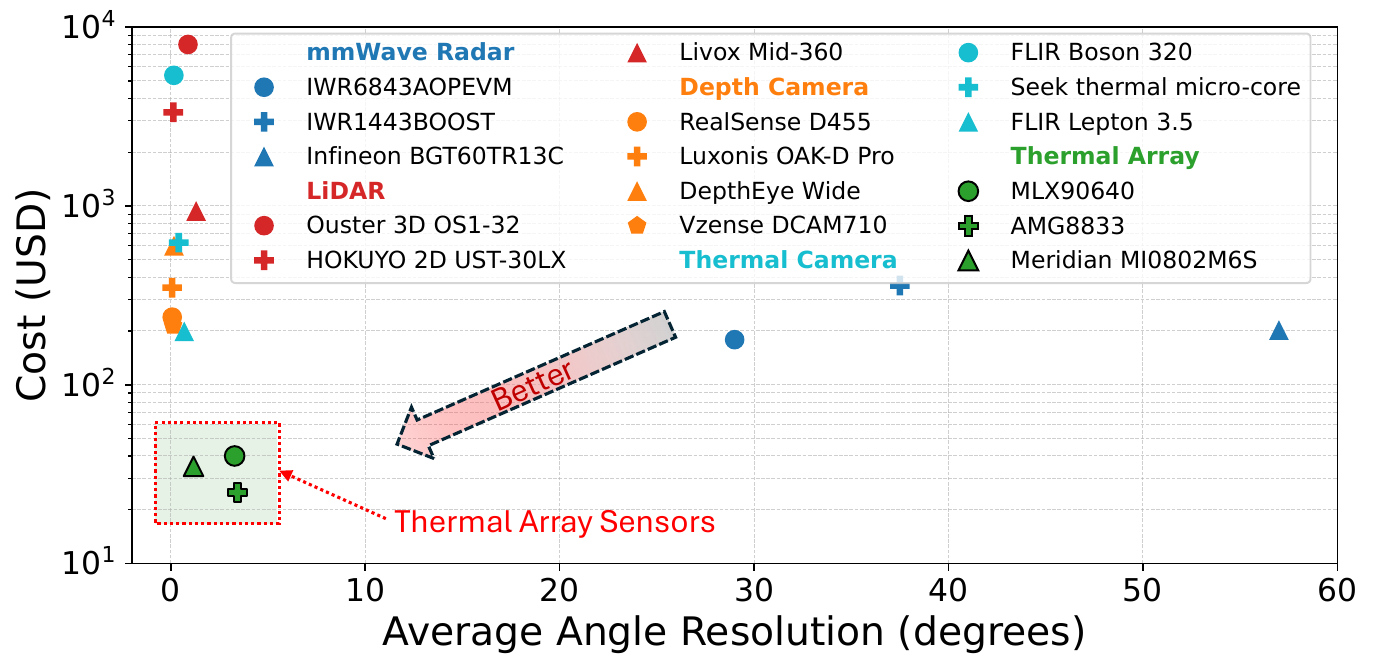}
  \vspace{-1.0\baselineskip}
  \caption{Cost versus angular resolution across different sensing modalities. 
  \rm{Thermal array sensors achieve a favorable balance between spatial resolution, cost, and privacy.}
  }
\label{fig:cost_angle}
\end{figure}

Recently, thermal arrays have emerged as a promising modality for human sensing \cite{naserAdaptiveThermalSensor2021, zhangTADARThermalArraybased2024, huFingerTrakContinuous3D2020, Zhangtapor2025}. 
By capturing long-wavelength infrared (LWIR) emissions, they produce human body heat signatures in the form of temperature maps (\fig\ref{fig:modality_overview}(b)) and offer several distinctive advantages (\tab\ref{tab:modality_comparison}):   

\head{\ding{182} Cost–resolution balance}
Thermal arrays offer a favorable trade-off with higher imaging resolution than compact mmWave radars while lower cost than LiDAR or depth cameras (\fig\ref{fig:cost_angle}), promising dense point cloud generation at scale. 

\head{\ding{183} Long-wavelength sensing}
By directly capturing heat signatures, thermal sensing inherently preserves privacy, achieves high sensitivity to human presence without the complex target-isolation processing of LiDAR or radar, and remains robust in visually challenging conditions (e.g., smoke, fog, or low light) where optical sensors fail \cite{kutukSemanticSegmentationThermal2022, liuSPECTRADronebasedMultispectral2024}.

\head{\ding{184} Fully passive operation}
Thermal arrays capture only body heat emissions without transmitting any signals, avoiding emission-related safety risks or cross-device interference. 

\noindent It is worth noting that while these advantages may appear modest in single-sensor scenarios, they become crucial in large-scale, ubiquitous deployments like smart buildings, where cost, cross-device interference, safety, and privacy concerns are amplified, rendering thermal sensing a superior choice over traditional alternatives.

\begin{figure*}[h]
\begin{minipage}{0.45\textwidth}
  \centering
  \includegraphics[width=0.99\textwidth]{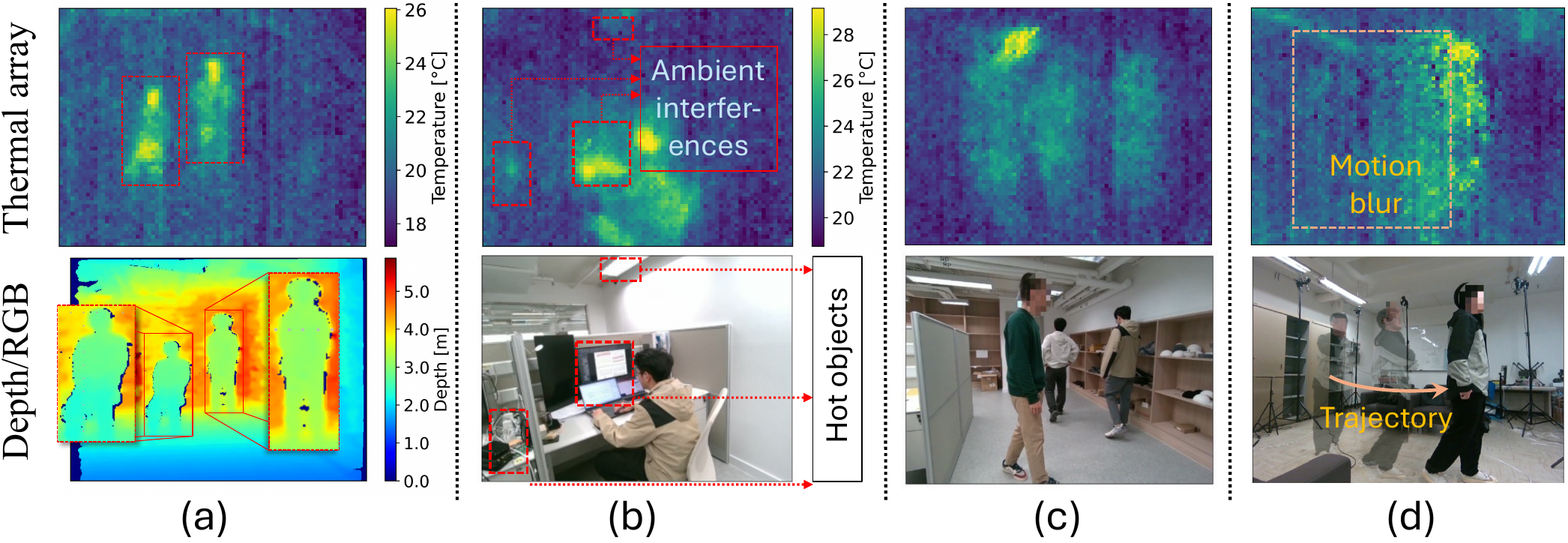}
  \vspace{-0.1\baselineskip}
  \caption{Challenges in generating point clouds from thermal data. \rm{Columns: (a) Depth estimation: thermal intensity is a poor proxy for geometry. (b) Thermal interference: ambient heat sources can cause false positives. (c-d) Multi-user ambiguity: adjacent signatures may merge and motion blur further obscures body boundaries.}}
\label{fig:challenge}
\end{minipage}
\hfill
\begin{minipage}{0.54\textwidth}
\centering
  \includegraphics[width=0.9\textwidth]{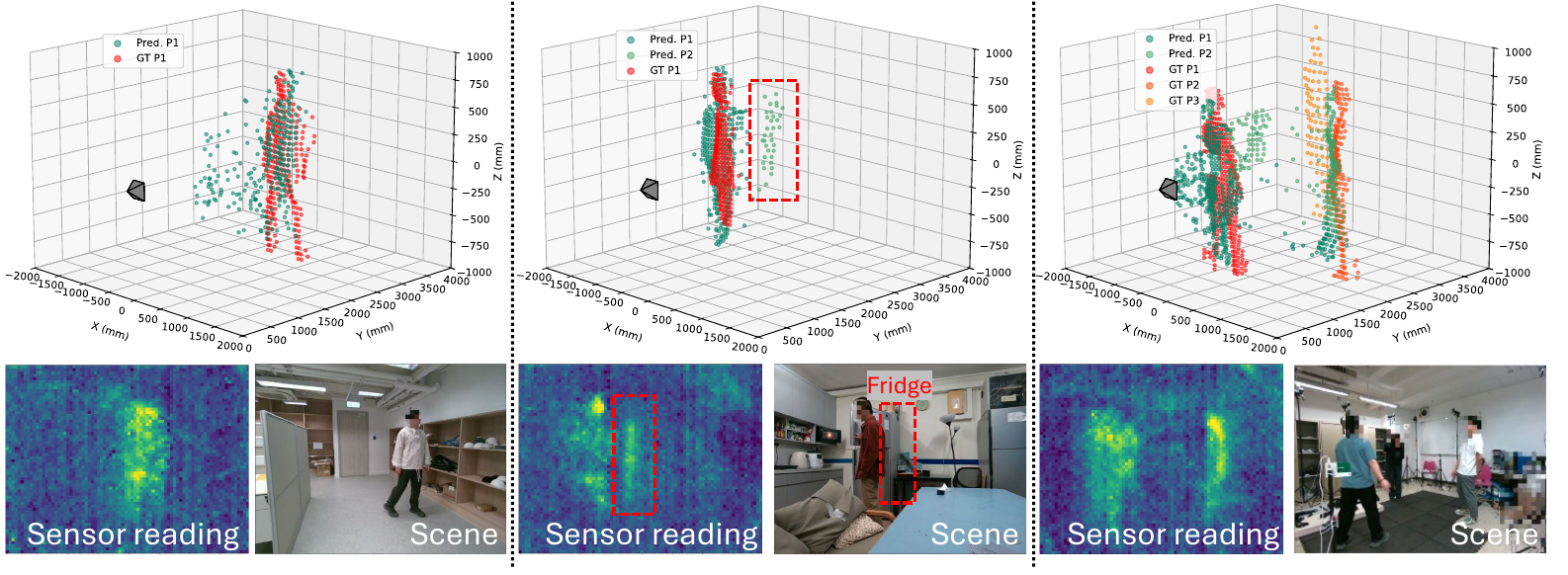}
  \vspace{-0.8\baselineskip}
  \caption{Qualitative results from the pilot study illustrating the key challenges. \rm{(Left) Depth estimation error, where the predicted (Pred.) point cloud deviates from the ground truth (GT). (Middle) Thermal interference, where a warm refrigerator (dotted red block) confuses the detection. (Right) Multi-user separation failure, where two individuals are incorrectly merged into one point cloud.}}
\label{fig:pilot_challenges_vis}
\end{minipage}
\end{figure*}

However, reconstructing dense human point clouds with a commodity thermal array remains challenging (\S\ref{sec:pilot}).

\head{$\bullet$ Accurate depth estimation}
Estimating depth from a temperature map is difficult, as the relationship between measured temperature intensity and body geometry is highly non-linear.  
As shown in \fig\ref{fig:challenge}(a), the non-uniform mapping between temperature values and spatial geometry necessitates an advanced depth estimation approach.

\head{$\bullet$ Thermal interference suppression}
Extracting human signals from ambient heat is non-trivial.  
As shown in \fig\ref{fig:challenge}(b), electronics, appliances, or sunlit furniture can emit heat signatures comparable to the human body, complicating foreground–background discrimination.

\head{$\bullet$ Multi-person spatial disentanglement}
Distinguishing multiple individuals in a low-resolution, textureless temperature map is difficult.  
As illustrated in \fig\ref{fig:challenge}(c-d), close proximity and motion blur often merge adjacent signatures and obscure body boundaries, degrading per-person representation and impairing downstream tasks such as instance-level reconstruction or activity analysis.  

To address these challenges and enable Thermal-Assisted 3D Point cloud reconstruction for human sensing, we introduce \sysname, the first system that generates dense human body point clouds using a low-cost thermal array sensor.
Leveraging the physics of thermal radiation, \sysname reconstructs accurate per-user 3D point clouds (\fig\ref{fig:modality_overview}(c)) from 2D temperature maps (\fig\ref{fig:modality_overview}(b)) by deriving one 3D point per temperature pixel. 
Although sparser than typical RGB-D outputs, this representation is substantially denser than those from compact radar or prior thermal ranging systems like TADAR \cite{zhangTADARThermalArraybased2024}, providing sufficient detail for fine-grained tasks such as 3D tracking and mesh reconstruction (\S\ref{sec:case}).
At the core of \sysname is a physics-informed design that integrates thermal physics with deep learning, optimized in an end-to-end manner. 
Specifically, the framework begins by modeling the relationship between human body geometry and captured temperature intensity.
By constructing a \textit{forward physics model} of thermal emission, propagation, and reception, this approach formulates the \textit{inverse problem} of point cloud reconstruction from thermal measurements.
From this formulation, \sysname integrates two key modules:

\head{$\blacksquare$ Multi-primitive estimation}
Our forward physics model reveals the fundamental difficulty of depth estimation: 
depth is entangled with other physical primitives such as emissivity and surface temperature.  
To address this, we introduce a \textit{Multi-Primitive Estimation Module}, a self-supervised network that implements an inverse problem solver for joint estimation of depth, emissivity, surface temperature, and reflection from the temperature map, thereby disentangling these primitives.  
The forward physics model then reconstructs the temperature map from these estimates, enabling physics-guided representation learning through self-supervised training without requiring ground-truth primitive values that are impractical to obtain in real-world scenarios.  

\head{$\blacksquare$ Geometric perspective fusion}
To address thermal interference and multi-person separation, we propose the \textit{Geometric Perspective Fusion Module}, which operates on the estimated primitives from two orthogonal perspectives.  
For interference suppression, we leverage the key insight that surface temperature and emissivity are distinctive human features.
Human detection is therefore performed on multi-channel Optical-Axis-View (OAV) maps comprising depth, emissivity, and surface temperature.  
For multi-person separation, we exploit the observation that individuals who appear partially overlapping in OAV can often be distinguished from a Bird’s-Eye View (BEV).
To achieve this, we propose Differentiable Index Mapping (DIM), which transforms OAV depth maps into BEV representations based on the sensor’s FoV and resolution.
This differentiable transformation is integrated into the network, supporting end-to-end OAV-to-BEV conversion.  
Finally, results from both perspectives are fused to produce refined per-person point clouds.

To evaluate the effectiveness of \sysname, we construct a large-scale dataset of over 160,000 thermal array samples collected across eight indoor environments with 11 participants.
Experimental results show that \sysname achieves high geometric accuracy in point cloud reconstruction with a mean directed Chamfer distance of 4.60 cm and a mean F1 score of 0.95 across both single- and multi-user cases. 
We further validate the practical utility of the reconstructed point clouds via three downstream tasks, all achieving remarkable performance: fall detection (91.46\% accuracy), indoor human tracking (mean absolute error of 21.86 cm), and human mesh recovery (a minimum matching distance of 4.87 cm). 
To the best of our knowledge, \sysname is the first system to transform body heat signatures into human point clouds, advancing privacy-preserving and fully passive human sensing while underpinning multi-sensor and cross-modality fusion with LiDAR, radar, depth cameras, and other modalities.

\head{Contributions} Our core contributions are as follows:

\begin{figure*}[t]
  \includegraphics[width=0.9\textwidth]{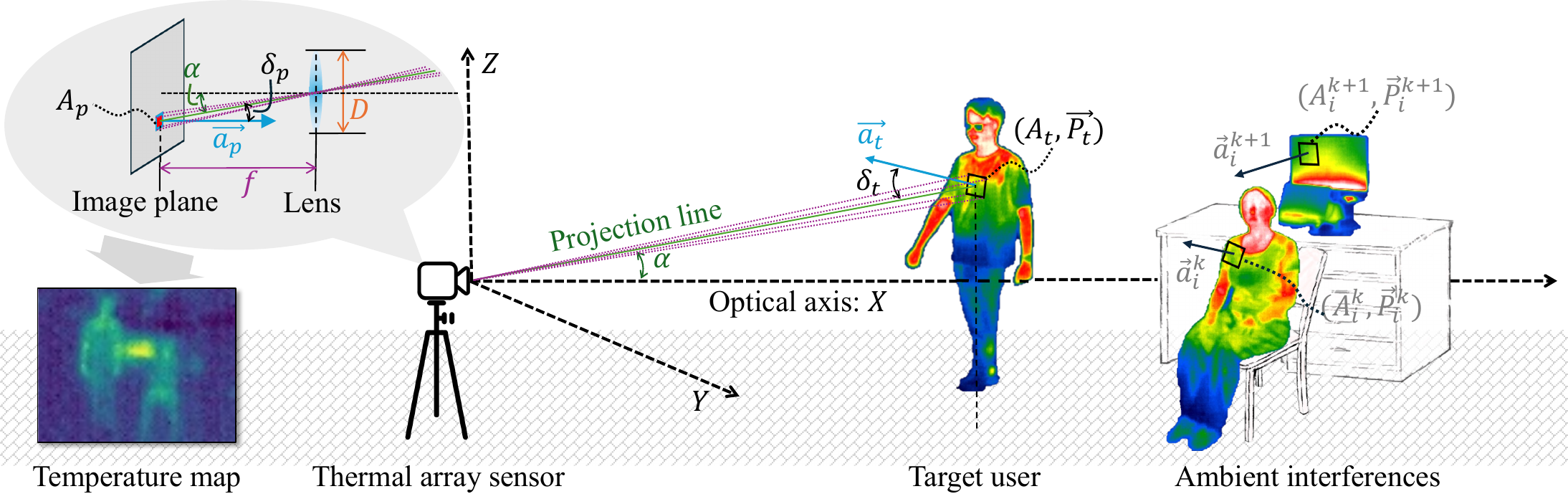}
  \vspace{-0.8\baselineskip}
  \caption{%
      Physical model of thermal array-based sensing. 
      \rm{Each sensor pixel \( A_p \) receives radiation from a target surface region \( A_t \), characterized by position \( \vec{P}_t \) and normal \( \vec{a}_t \). 
      Radiation undergoes atmospheric attenuation and may include reflections from ambient interferences \( A_i^k \). 
      Projection follows the pinhole camera model with parameters \( f \), \( D \), \( \alpha \), and \( \delta_t \), yielding the final temperature map.}
      }
\label{fig:theory_diagram}
\end{figure*}

\begin{itemize}[noitemsep, topsep=0pt, leftmargin=*]
\item We present \sysname, the first system that taps into low-cost thermal array data to reconstruct dense 3D human point clouds, enabling a new paradigm for privacy-preserving, fully passive human sensing applications. 

\item We propose physics-informed multi-primitive estimation, which leverages a forward thermal physics model to self-supervise an inverse problem solver for jointly recovering depth and related thermal properties.
We also present novel geometric perspective fusion, introducing Differentiable Index Mapping for OAV-to-BEV transformation to separate users and suppress ambient interference. 

\item We curate a large-scale dataset and validate \sysname through extensive evaluation and real-world case studies, demonstrating consistently strong performance. We fully open-source \sysname at \href{https://github.com/aiot-lab/TAP3D}{\color{blue}\texttt{https://github.com/aiot-lab/TAP3D}}.
\end{itemize}

\section{Pilot Study}
\label{sec:pilot}
To empirically ground the challenges outlined in \S\ref{sec:intro}, we conduct a pilot study that demonstrates the non-trivial nature of reconstructing high-fidelity human point clouds from a temperature map. 
We develop a baseline deep learning model and evaluate its performance on a self-collected dataset. 
The results reveal the practical limitations of a straightforward learning approach and motivate the necessity for the advanced techniques developed in \sysname.

\head{Data collection}
We collect data from five volunteers in five distinct indoor environments using a thermal array sensor.
The dataset contains 36,921 temperature maps with ground truth point clouds and masks, obtained from a co-located Intel RealSense D455 depth camera.
We split the data into 17,956 training and 18,965 test samples, covering single-user, multi-user, and background-only scenarios.

\begin{table}[b]
\caption{Pilot study results.}
\centering
\label{tab:pilot_results}
\begin{tabular}{lccc}
\hline
Scenario      & DCD (cm) $\downarrow$ & F1 Score $\uparrow$ & MAE (cm) $\downarrow$ \\ \hline
Overall       & 22.6        & 0.88       & 31.94        \\
Single-User   & 18.5        & 0.91       & 26.83        \\
Multi-User    & 34.2        & 0.82       & 51.72        \\ \hline
\end{tabular}
\end{table}

\head{Baseline model}
We implement a U-Net-like baseline model \cite{ronnebergerUNetConvolutionalNetworks2015} that takes a single-channel temperature map as input and outputs three maps:
(1) a dense depth map, (2) an instance segmentation map assigning unique IDs to each person based on distance, and (3) a binary foreground–background mask.
The final 3D point cloud is generated by projecting the segmented depth into 3D space using the predicted depth and the sensor’s intrinsic parameters.

\head{Results and analysis}
We evaluate the baseline model using standard metrics for point cloud reconstruction. 
\tab\ref{tab:pilot_results} summarizes the results across all test data, including both single-user and more challenging multi-user scenarios. 
While the model achieves a reasonable F1 score of 0.91 in single-user detection, performance drops markedly in multi-user scenes, with the F1 score decreasing to 0.82.
In terms of depth estimation, the mean absolute error (MAE) increases from 26.83 cm in single-user settings to 51.72 cm in multi-user cases. More critically, geometric accuracy, measured by the Directed Chamfer Distance (DCD), deteriorates from 18.5 cm to 34.2 cm, reflecting the model’s difficulty in handling inter-person occlusion and separation.
These findings confirm that a straightforward learning approach is insufficient and highlight the need for a principled framework, as further illustrated by the qualitative results in \fig\ref{fig:pilot_challenges_vis}.

\begin{figure*}[t]
  \includegraphics[width=0.95\textwidth]{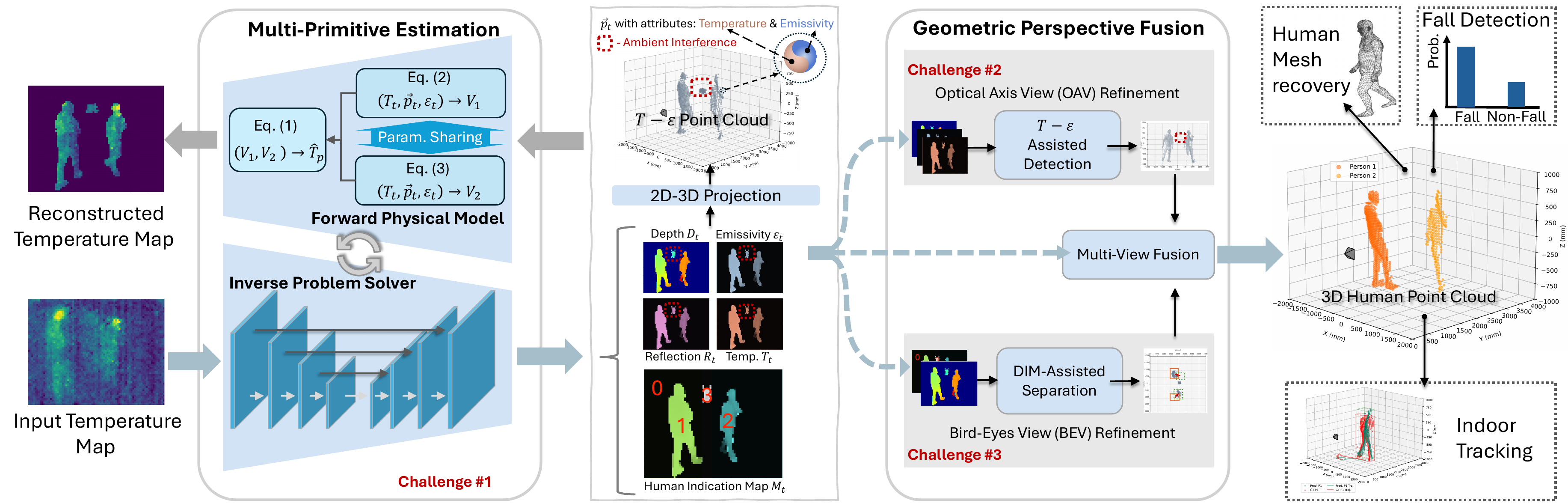}
  \caption{\sysname model structure. \rm{It consists of a Multi-Primitive Estimation Module that jointly estimates depth, emissivity, temperature, and reflection under physics-based self-supervision, and a Geometric Perspective Fusion Module that refines results through OAV (interference suppression) and BEV (multi-user separation), yielding accurate per-person 3D point clouds.}}
\label{fig:model_structure}
\end{figure*}

\section{\sysname Physics Model}
\label{sec:physics}

We establish a physics model that links human-body primitives (3D location, surface temperature, and emissivity) to the temperature values reported by a thermal array sensor, grounded in thermography theory \cite{mollmannInfraredThermalImaging2018}.  
As shown in \fig\ref{fig:theory_diagram}, this forward model describes how body-emitted and reflected thermal radiation propagates through the atmosphere, passes the optics, and is converted into pixel-wise temperatures, and forms the basis of the inverse problem of recovering 3D human point clouds from thermal measurements.  
A detailed derivation is given in Appendix~\ref{app:physics_model}; here we summarize the resulting inverse formulation used by \sysname.

We group all variables into three sets: (1) unknown target-related primitives: spatial location $\overrightarrow{P}_t$, surface temperature $T_t$, and emissivity $\epsilon_t$; (2) fixed parameters: ambient constants $\overrightarrow{P}_i^k$, $M_{\lambda,k}$, $\xi_t$, device parameters $G$, $D$, $f$, $A_p$, and sensor constant $\zeta = N \kappa \rho / G_{\text{th}}$, and environmental attenuation $\gamma$; and (3) observed quantities: the sensor-estimated temperature $\hat{T}_p$.
The inverse problem of point cloud estimation is formulated as recovering $\overrightarrow{P}_t$ from $\hat{T}_p$ via the Sakuma--Hattori model:
\begin{equation}
\hat{T}_p\!\left(\overrightarrow{P_t}; T_t, \epsilon_t \right)
= \frac{C_2}{C_3 \ln{\left(\frac{C_1}{V_1 + V_2} + 1\right)}} + \frac{C_4}{C_3},
\label{eq:physic_final}
\end{equation}
where $T_t$ and $\epsilon_t$ are human-body primitives, $V_1$ and $V_2$ are the voltage signals from self-emission and reflected ambient radiation, and $C_1$--$C_4$ are sensor-specific calibration coefficients.
The self-emission term $V_1$ is
\begin{equation}
V_1 = \frac{G D^2 \zeta A_p}{4f^2} \,
    \frac{\langle \vec{P_t}, \vec{x} \rangle^4 e^{-\gamma \|\vec{P_t}\|}}{\|\vec{P_t}\|^4}
    \, \epsilon_t \int_{\lambda_1}^{\lambda_2} M_{\lambda}^{\text{BB}}(T_t) \, d\lambda,
\label{eq:physic_v1}
\end{equation}
and the reflected term $V_2$ is
\begin{equation}
V_2 = \frac{G D^2 A_p \xi_t \zeta}{4f^4} \,
    \frac{\langle \vec{P_t}, \vec{x} \rangle^7 e^{-\gamma \|\vec{P_t}\|}}{\|\vec{P_t}\|^5}
    \, (1 - \epsilon_t) \sum_{k=1}^{K} e^{-\gamma \|\vec{P}_t - \vec{P}_i^k\|} \, S_{\lambda,k},
\label{eq:physic_v2}
\end{equation}
where $\vec{x}$ is the unit vector along the optical axis and
$S_{\lambda,k} = \int_{\lambda_1}^{\lambda_2} M_{\lambda,k} \, d\lambda$ is the integrated spectral exitance of the $k$-th ambient region.
Equations~\eqref{eq:physic_final}–\eqref{eq:physic_v2} define the physics-constrained inverse problem whose solution yields dense human point clouds from thermal array measurements.

\section{Physics-Informed Design}
\label{sec:co_desgin_model}

In this section, we present the \sysname model for reconstructing 3D human point cloud from a temperature map. 
This model adopts a physics-informed design framework that integrates the physical model developed in \S\ref{sec:physics} with a data-driven approach to address the inverse problem.
We begin by providing an overview of the model architecture, followed by a detailed description of the key components that address the specific challenges outlined in \S\ref{sec:pilot}.

\subsection{\sysname Model Overview}
\label{subsec:model_overview}
\sysname reconstructs accurate 3D human point clouds from a \textit{single}  temperature map captured by a thermal array sensor.
This capability supports a wide range of downstream tasks and enables multi-sensor and cross-modality sensing. 
As illustrated in \fig\ref{fig:model_structure}, \sysname consists of two main components:  
(1) \textbf{Multi-Primitive Estimation Module} for accurate depth recovery, addressing the first challenge in \S\ref{sec:pilot}; and  
(2) \textbf{Geometric Perspective Fusion Module} that refines the estimated primitives to suppress ambient interference and disentangle multiple users.  
The refined outputs from both views are fused to produce accurate 3D human point clouds for applications such as mesh recovery, fall detection, and indoor tracking.

\subsection{Multi-Primitive Estimation}
\label{subsec:multi-primitive}

We propose a multi-primitive estimation approach for accurate depth recovery.

\head{Root cause of depth estimation difficulty}  
As established in \S\ref{sec:physics}, depth (\(\langle \vec{P_t}, \vec{x} \rangle\) in \eqn\eqref{eq:physic_v1} and \eqn\eqref{eq:physic_v2}) is nonlinearly coupled with other human-body primitives, including emissivity \(\epsilon_t\) and surface temperature \(T_t\). 
This coupling creates a non-uniform correspondence between the measured temperature \(\hat{T}_p\) and the depth. 
Such entanglement poses a fundamental challenge for accurate depth estimation.  
To resolve this issue, we integrate an \textit{inverse problem solver} with the \textit{forward physical model}, enabling self-supervised estimation of all relevant primitives and thereby disentangling depth from emissivity and surface temperature.

\head{Inverse problem solver}  
As discussed in \S\ref{sec:physics}, estimating human-body primitives from thermal measurements is an ill-posed inverse problem. 
We adopt a U-Net–like neural network as the inverse problem solver to address this challenge in a data-driven manner.
Given a temperature map as input, the network jointly estimates depth $D_t$, emissivity $\epsilon_t$, reflected radiation $R_t$, and surface temperature $T_t$ for all candidate regions. 
To distinguish different regions, the inverse solver additionally outputs a human indication map $M_t$, where pixels with value 0 denote non-human background, and pixels with values 1, 2, \dots\ correspond to human instances (user\#1, user\#2, \dots).
The estimated depth map is then converted into a 3D point cloud based on the sensor’s FoV and resolution. 
\textit{Since ground truth for emissivity, surface temperature, and reflected radiation is unavailable in practical non-contact scenarios, a forward physical model is introduced to provide supervisory signals.}

\begin{figure}[t]
  \includegraphics[width=0.46\textwidth]{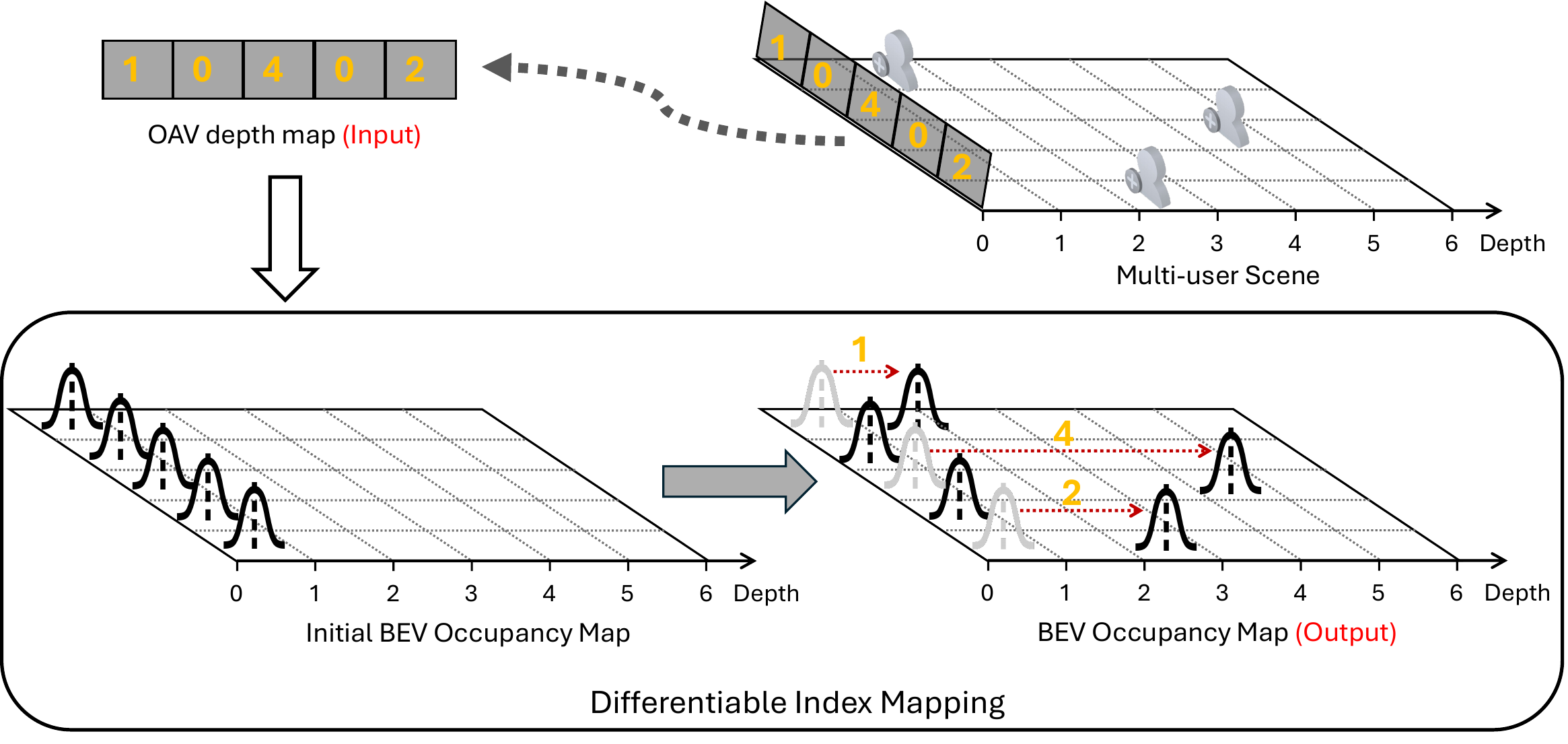}
  \vspace{-0.8\baselineskip}
  \caption{Differentiable Index Mapping (DIM). \rm{The DIM operation converts a depth map in OAV into a BEV occupancy map in a differentiable and invertible manner.}}
\label{fig:dim}
\end{figure}

\head{Forward physical model}  
The forward physical model implements the formulation in \eqn\eqref{eq:physic_final}–\eqref{eq:physic_v2}, taking the estimated primitive maps as input to reconstruct the corresponding temperature map $T_{recon}$.
To keep the model computationally efficient and tractable, we adopt two approximations:  
(1) The spectral integration in Planck’s law, $\int_{\lambda_1}^{\lambda_2} M_{\lambda}^{\text{BB}}(T_t)\, d\lambda$ in \eqn\eqref{eq:physic_v1}, is approximated by the Stefan–Boltzmann law, $\sigma T_t^4$, where $\sigma \approx 5.670 \times 10^{-8} \ \text{W}\,\text{m}^{-2}\,\text{K}^{-4}$.  
(2) The high-order reflected radiation term, $\sum_{k=1}^{K} e^{-\gamma \|\vec{P_t} - \vec{P}_i^k\|} \cdot S_{\lambda,k}$ in \eqn\eqref{eq:physic_v2}, is replaced by a reflected-radiation map directly predicted by the inverse problem solver.  
All other parameters are treated as learnable and optimized jointly with the network.  

\head{Physics-guided representation learning} 
Coupling the inverse problem solver with the forward physical model establishes a self-supervised training scheme. 
The network parameters are optimized by minimizing the reconstruction loss between the forward model output and the input temperature map, enabling primitive estimation without explicit ground truth for emissivity, surface temperature, and reflected radiation.
This physics-guided learning approach significantly improves label efficiency and reconstruction robustness, laying the groundwork for fully self-supervised thermal reconstruction in future research.

\subsection{Geometric Perspective Fusion}
\label{subsec:multi-view}
To enhance reconstruction quality, we introduce a Geometric Perspective Fusion Module that processes the estimated multi-primitive maps from two complementary perspectives: the \textit{optical-axis view} (OAV) and the \textit{bird’s-eye view} (BEV). 
This design exploits complementary spatial cues, enabling robust interference suppression and multi-user separation.

\head{OAV-based refinement} 
In the OAV, ambient thermal interference is suppressed by leveraging physical cues unique to the human body. 
Specifically, we jointly consider emissivity \(\epsilon_t\) and surface temperature \(T_t\) as primary discriminative features, combined with depth \(D_t\) to form a three-channel input.
These inputs are fed into a lightweight convolutional network, termed the \textit{\(T\)–\(\epsilon\) assisted detection model}, which outputs a binary mask \(M_{\text{OAV}} \in \mathbb{R}^{K \times I}\), where pixels labeled as 1 correspond to human regions.

\head{BEV-based refinement}
For multi-user separation, we operate in the BEV space to disentangle subjects more effectively.
Rather than hallucinating fully occluded regions, this conversion targets separation and depth refinement by mapping partially overlapping or connected OAV signatures into distinct BEV locations.
Since the estimated primitives are represented in OAV, a differentiable and invertible transformation is required for end-to-end training. 
To this end, we propose \textit{Differentiable Index Mapping} (DIM), illustrated in \fig\ref{fig:dim}.
Given an OAV depth map $D \in \mathbb{R}^{K \times I}$, where $K$ is the depth map height and $I$ is the width, DIM projects each pixel depth value $d_{k,i}$ into the BEV coordinate frame as a Gaussian distribution over horizontal positions.  
Formally, the BEV occupancy map $O \in \mathbb{R}^{I \times J}$ is computed as:
$
O(i,j) = \sum_{k} \mathcal{N}(j \mid d_{k,i}, \sigma),
$
where $\mathcal{N}(j \mid d_{k,i}, \sigma)$ is a Gaussian kernel centered at $d_{k,i}$, and $j$ indexes discretized depth bins, yielding a 2D spatial grid spanned by the horizontal axis $I$ and $J$ depth bins.
Importantly, DIM is invertible, allowing the BEV occupancy map to be mapped back to the OAV depth domain with arbitrarily fine reconstruction precision controlled by the Gaussian kernel~$\sigma$.
The \textit{DIM-assisted separation model} then works as follows:
the OAV depth map \(D_t\) is transformed into BEV via DIM, \ie, \(O_t = \text{DIM}(D_t)\), and a lightweight neural network refines it to produce the BEV occupancy map \(O_{\text{BEV}}\).

\head{Multi-view fusion}  
We merge the OAV and BEV outputs with the original depth map from the multi-primitive estimation module to produce the final human point cloud.  
From the BEV occupancy map $O_{\text{BEV}} \in \mathbb{R}^{I \times J}$, we derive per-column depth bounds in OAV space:  
\begin{align}
D_{\min}(i) &= \min\{\, j \mid O_{\text{BEV}}(i,j) = 1 \,\}, \\
D_{\max}(i) &= \max\{\, j \mid O_{\text{BEV}}(i,j) = 1 \,\}.
\end{align}
The refined depth map $D_t^{*} \in \mathbb{R}^{K \times I}$ is obtained by masking $D_t$ with the OAV segmentation $M_{\text{OAV}}$ and clamping each column to the valid range:
\begin{equation}
D_t^{*}(k,i) =
\begin{cases}
\text{Clip}\big(D_t(k,i), D_{\min}(i), D_{\max}(i)\big), & M_{\text{OAV}}(k,i) = 1, \\
0, & \text{otherwise},
\end{cases}
\end{equation}
where $\text{Clip}(x,a,b) = \min(\max(x,a),b)$.
Finally, the output point cloud is generated by converting the refined depth map into 3D coordinates using the sensor’s FoV and resolution.

\subsection{Training Strategy}
\label{subsec:training}
We train \sysname end-to-end with a weighted combination of self-supervised and supervised losses.
Ground-truth depth maps $D_g$ and human indication maps $M_g$ are obtained from a co-located RealSense D455.
The multi-primitive estimation module is optimized with Huber losses on temperature reconstruction, depth, and indication maps, while the geometric perspective fusion module uses binary cross-entropy losses for OAV segmentation and BEV refinement.
All terms are combined as $\mathcal{L} = \sum_{i=1}^{5} \lambda_i L_i$, where $\lambda_i$ are weighting factors and the full loss definitions are provided in Appendix~\ref{app:training}.

\begin{figure}[b]
  \includegraphics[width=0.46\textwidth]{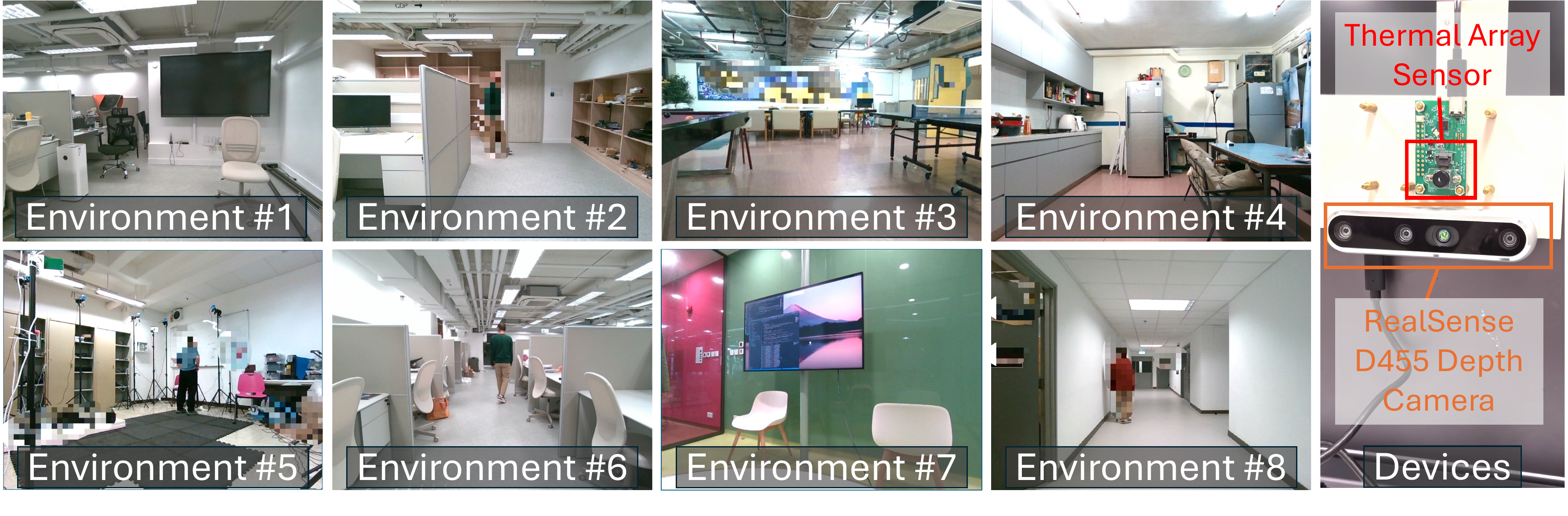}
  \vspace{-0.8\baselineskip}
  \caption{Experimental setup. \rm{We conduct experiments in eight indoor environments using a  thermal array sensor and co-located depth camera for ground truth acquisition.}}
\label{fig:setup_vis}
\end{figure}

\section{Implementation}
\label{sec:impl}

\head{Hardware} 
\sysname is implemented using the low-cost Meridian MI0802M6S thermal array sensor, featuring an $80 \times 62$ element array with a $90^{\circ} \times 67^{\circ}$ FoV at a cost of \(\sim\$10\). 
A co-located Intel RealSense D455 camera (\fig\ref{fig:setup_vis}) acquires ground-truth 3D point clouds and target masks.
Data is recorded through a unified host program at 8 Hz. Each thermal frame is temporally paired with the nearest depth frame via host timestamps, followed by spatial alignment using our custom calibration tool.

\head{Software} 
\sysname is implemented in PyTorch with 23.29M parameters and trained on an RTX 4090 GPU.  
We use the AdamW optimizer \cite{loshchilov2017decoupled} with a learning rate of 0.001 and weight decay of 0.01.  
Models are trained for up to 100 epochs with an early stopping patience of 8 epochs. 
In the DIM, $J=400$ and $\sigma=0.5$ correspond to the D455's 8000\,mm maximum range partitioned into 20\,mm depth bins.
Code, models, and datasets are at \href{https://github.com/aiot-lab/TAP3D}{\color{blue}\texttt{https://github.com/aiot-lab/TAP3D}}.

\begin{figure*}[t]
  \includegraphics[width=0.9\textwidth]{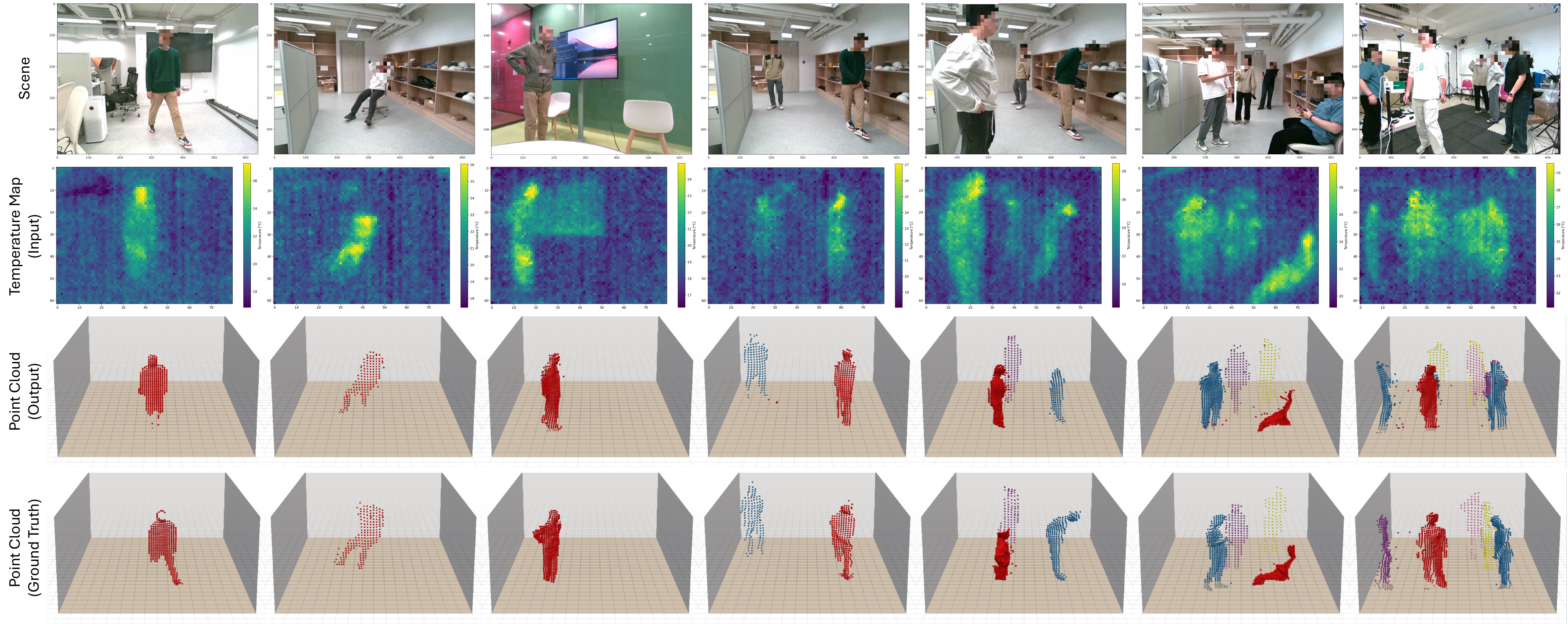}
  \vspace{-0.8\baselineskip}
  \caption{%
      Visualization of \sysname results. \rm{Top to bottom: (1) input scene, (2) temperature map from the thermal array sensor, (3) reconstructed human point clouds by \sysname, and (4) corresponding ground-truth point clouds.}
      }
\label{fig:output_vis}
\end{figure*}

\section{Experiments}
\label{sec:exp}

\subsection{Dataset, Baselines, and Metrics}
\label{subsec:dataset_baselines}

We overview the dataset, baselines, and metrics below, while providing full details in Appendix~\ref{app:dataset_baselines}.\\
\head{Dataset}
We evaluate \sysname on a large-scale dataset of 160{,}000+ synchronized thermal array samples collected in eight indoor environments from 11 volunteers (\fig\ref{fig:setup_vis}).
Each sample includes a temperature map, D455-based 3D point cloud ground truth, and a Detectron2 human mask \cite{wu2019detectron2}.
Data are recorded at 8 Hz in one-minute segments and split at the segment level into 88{,}631 training and 80{,}011 test samples. \\
\head{Baselines}
Since no direct thermal array-based point cloud baselines exist, we adapt an RGB-to-point cloud method, RGB2Point \cite{lee2025rgb2point}, and a thermal image-based depth estimation approach, NeWCRF \cite{yuan2022neural}, for thermal array inputs.
Rather than relying on off-the-shelf weights, we train all baseline models from scratch on the complete \sysname training split.
\\
\head{Metrics}
We report directed Chamfer distance (DCD) for geometric accuracy and F1 score for user detection.
For user detection, F1 is computed from precision and recall over bounding boxes extracted from predicted and ground-truth human indication maps at an IoU threshold of 0.5.

\begin{table}[t]
  \centering
  \footnotesize
  \caption{Comparison with SOTA baselines. \rm{\sysname achieves the best reconstruction accuracy while being significantly more parameter- and computation-efficient.}}
  \label{tab:model_compare}
  \vspace{-1.0\baselineskip}
  \begin{tabular}{lccc}
    \toprule
    Metrics & RGB2Point~\cite{lee2025rgb2point} & NeWCRF~\cite{yuan2022neural} & \textbf{\sysname} \\
    \midrule
    DCD (cm)$\downarrow$   & 10.15 $\pm$ 0.77 & 19.26 $\pm$ 1.50 & \textbf{4.60 $\pm$ 0.25} \\
    F1 Score$\uparrow$          & 0.945           & 0.677            & \textbf{0.953} \\
    Params (M)$\downarrow$& 176.07          & 88.46            & \textbf{23.32} \\
    FLOPs (G)$\downarrow$ & 16.96           & 78.14            & \textbf{7.41} \\
    \bottomrule
  \end{tabular}
\end{table}

\subsection{Overall Performance}
As illustrated in \fig\ref{fig:output_vis}, \sysname reconstructs dense human body point clouds across both single- and multi-user scenarios, demonstrating strong potential for diverse sensing applications and cross-modal integration. We next present a quantitative evaluation of its performance.

\head{Performance comparison with CV methods}
As summarized in \tab\ref{tab:model_compare}, \sysname consistently outperforms CV-based SOTA baselines in both reconstruction accuracy and efficiency. It achieves the lowest DCD (4.60\,cm vs.\ 10.15\,cm for RGB2Point and 19.26\,cm for NeWCRF) and highest F1 score (0.953), while requiring substantially fewer parameters (23.32 M) and lower computational cost (7.41 G FLOPs).
These results clearly demonstrate the advantage of our physics-informed, thermal array–specific design over direct image-to-3D or depth-estimation approaches.

\begin{figure}[h]
  \centering
  \includegraphics[width=0.48\textwidth]{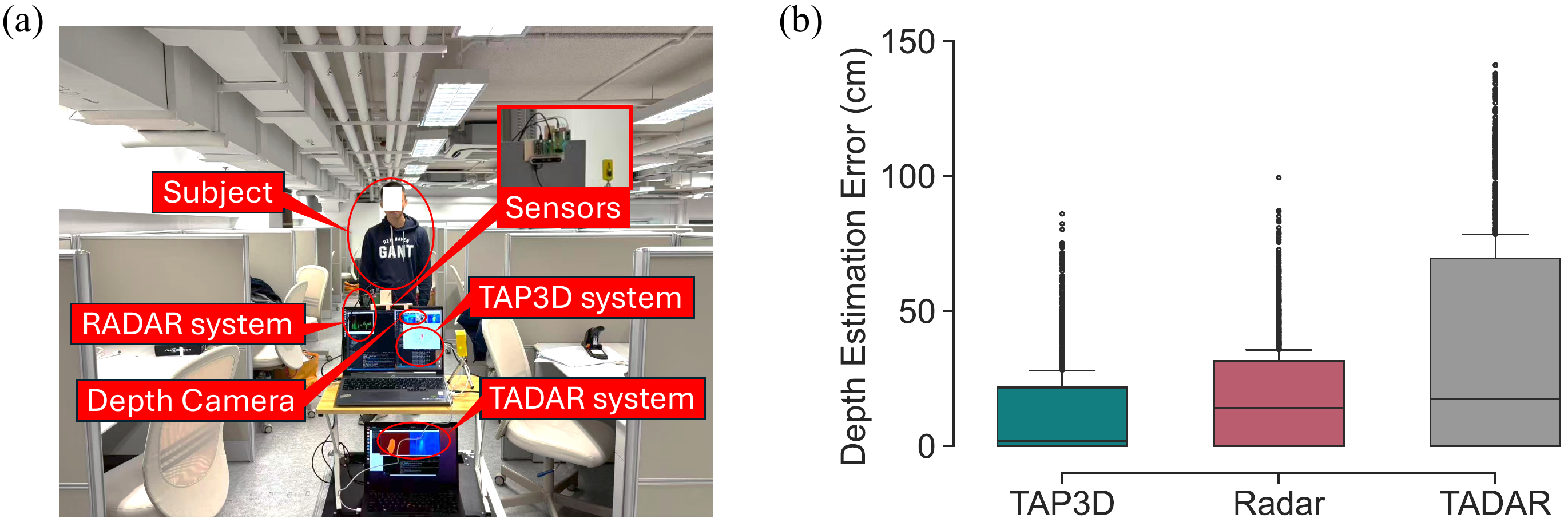}
  \vspace{-1.5\baselineskip}
  \caption{Depth estimation comparison between \sysname and other low-cost sensing systems. 
  \rm{(a) Experimental setup featuring the synchronized \sysname, TADAR \cite{zhangTADARThermalArraybased2024}, and compact radar (Infineon-BGT60TR13C) systems, with a RealSense D455 camera for ground truth. (b) Distribution of depth estimation errors.}
  }
\label{fig:system_compare}
\end{figure}

\head{Performance comparison with low-cost systems}
To evaluate the advantages of \sysname, we compare its \textit{depth} estimation accuracy with two representative low-cost alternatives: TADAR \cite{zhangTADARThermalArraybased2024} and a compact Infineon BGT60TR13C radar (1 Tx, 3 Rx), as shown in \fig\ref{fig:system_compare}(a).
While \sysname reconstructs per-pixel 3D point clouds, TADAR yields a single-depth human mask and the radar reports only a dominant reflection depth.
Therefore, we benchmark all systems using a single subject walking between 0.5\,m and 4\,m, with a RealSense D455 providing depth ground truth.
As shown in \fig\ref{fig:system_compare}(b), across 985 synchronized samples, \sysname achieves the lowest mean error of $13.04$\,cm.
The radar yields a higher error of $19.10$\,cm because its limited resolution captures only a single dominant reflection—often misaligned with the ground-truth depth averaged across all body pixels.
TADAR achieves a mean error of $36.74$\,cm, matching its reported range on the scale of human body thickness.
These results highlight that \sysname delivers superior depth precision and spatial fidelity compared to existing low-cost thermal and radar systems.

\begin{figure*}[t]
  \begin{minipage}{0.23\textwidth}
  \centering
  \includegraphics[width=\linewidth]{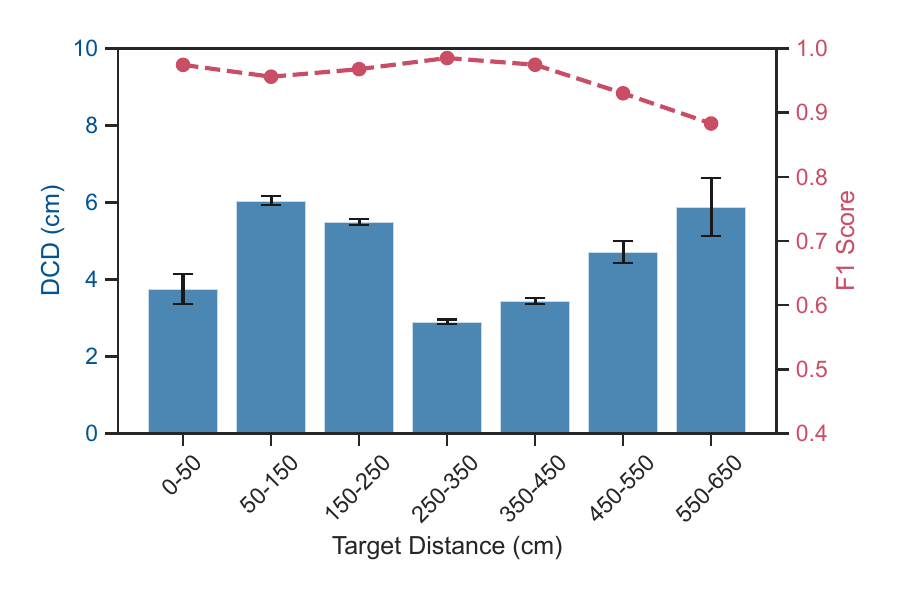}
  \vspace{-2\baselineskip}
  \caption{Overall performance.}
  \label{fig:overall_performance_distance_f1}
\end{minipage}
\hfill
\begin{minipage}{0.23\textwidth}
  \centering
  \includegraphics[width=0.99\textwidth]{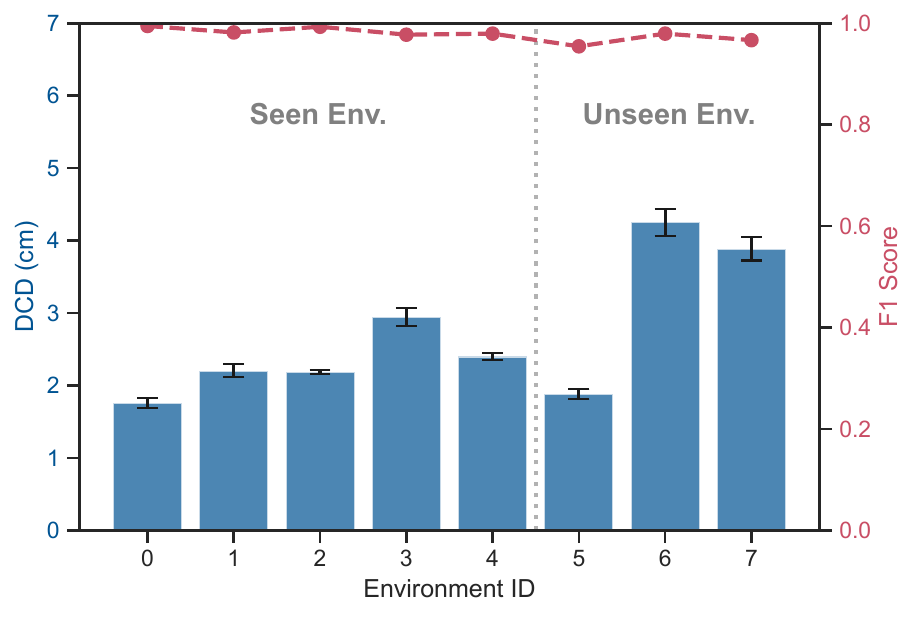}
  \vspace{-2\baselineskip}
  \caption{Results across environments.}
  \label{fig:cross_env}
  \end{minipage}
  \hfill
    \begin{minipage}{0.23\textwidth}
  \centering
 \includegraphics[width=0.99\textwidth]{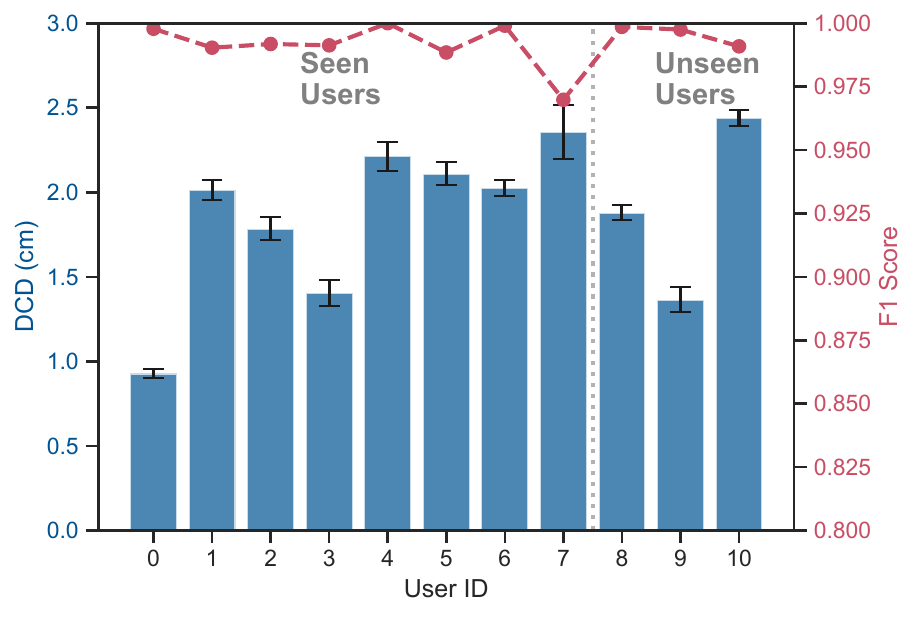}
  \vspace{-2\baselineskip}
  \caption{Cross-user performance.}
  \label{fig:cross_user}
\end{minipage}
\hfill
\begin{minipage}{0.23\textwidth}
  \centering
  \includegraphics[width=0.99\textwidth]{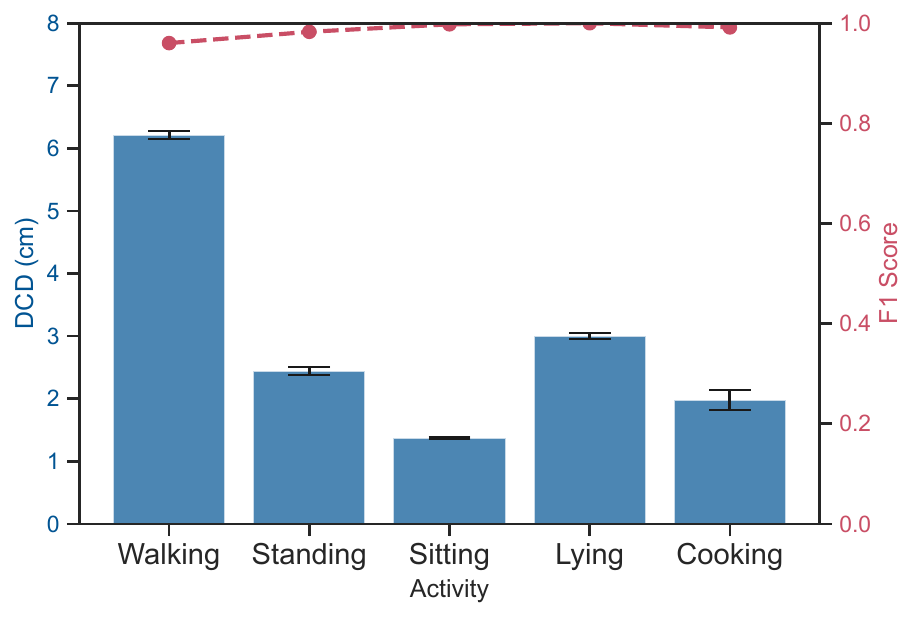}
  \vspace{-2\baselineskip}
  \caption{Impact of activity.}
  \label{fig:cross_activity}
\end{minipage}
\end{figure*}

\begin{figure*}[t]
\begin{minipage}{0.23\textwidth}
  \centering
  \includegraphics[width=0.99\textwidth]{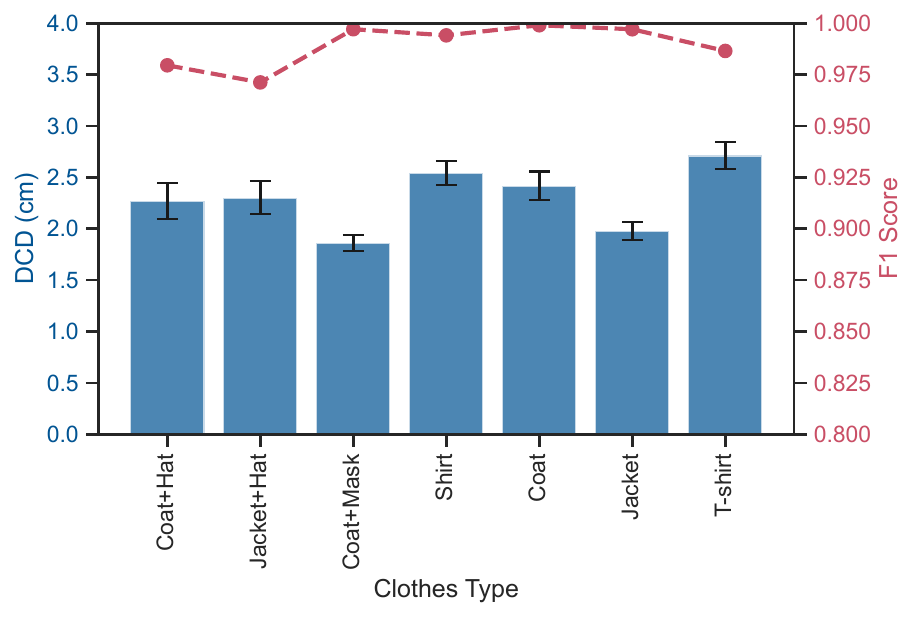}
  \vspace{-2\baselineskip}
  \caption{Impact of clothing.}
  \label{fig:cross_clothes}
\end{minipage}
\hfill
\begin{minipage}{0.23\textwidth}
  \centering
 \includegraphics[width=0.99\textwidth]{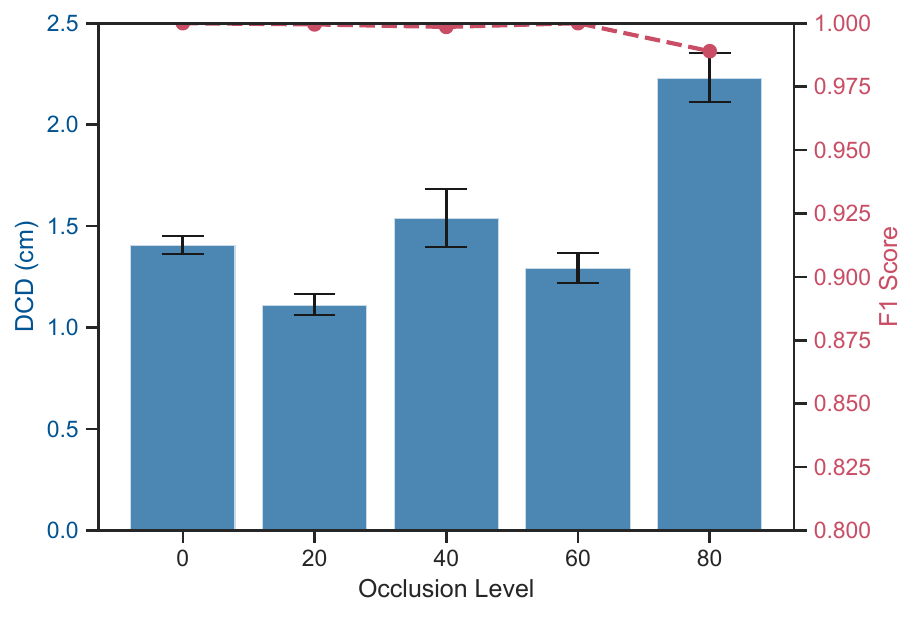}
  \vspace{-2\baselineskip}
  \caption{Impact of the occlusion. }
  \label{fig:cross_occlusion}
\end{minipage}
\hfill
\begin{minipage}{0.23\textwidth}
  \centering
  \includegraphics[width=0.99\textwidth]{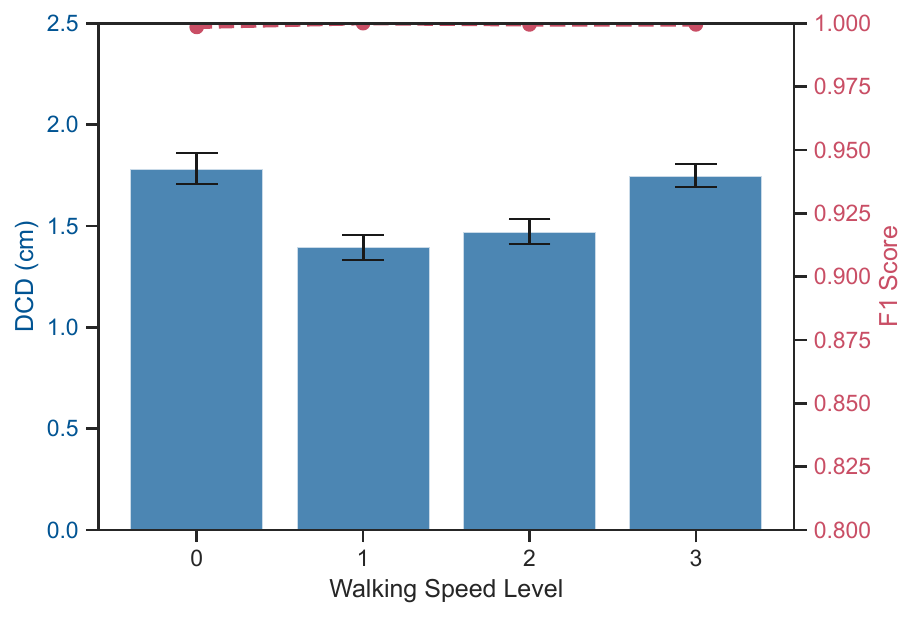}
  \vspace{-2\baselineskip}
  \caption{Walking speed robustness.}
  \label{fig:cross_walking_speed}
  \end{minipage}
\hfill
\begin{minipage}{0.23\textwidth}
  \centering
  \includegraphics[width=0.99\textwidth]{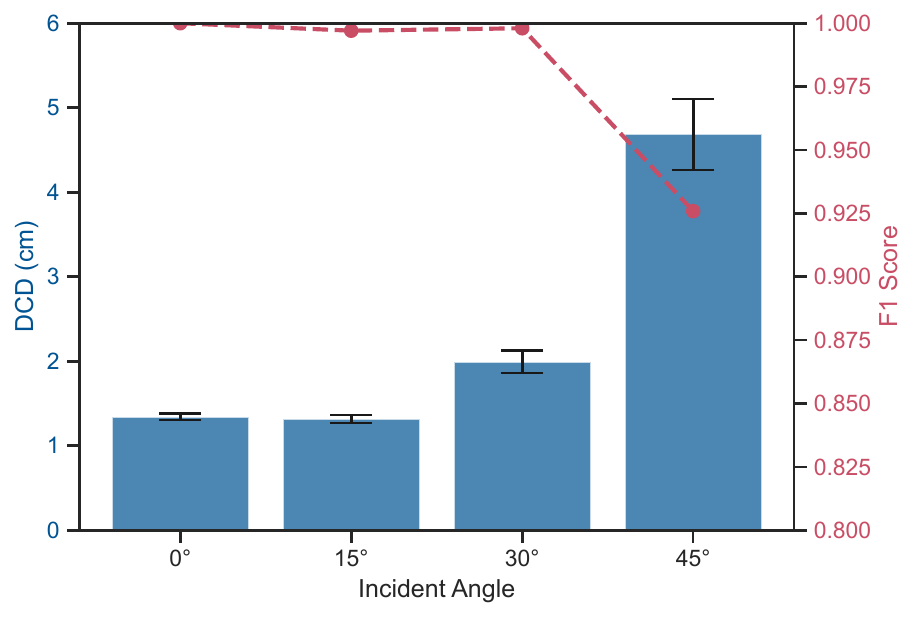}
  \vspace{-2\baselineskip}
  \caption{Incident angle impact.}
  \label{fig:cross_incident_angle}
\end{minipage}
\end{figure*}

\head{Details on detection and geometric reconstruction}  
As shown in \fig\ref{fig:overall_performance_distance_f1}, \sysname achieves robust performance up to 6.5 m, a sufficient range for typical indoor applications.
Although the F1 score decreases slightly with distance, it remains near 0.90 at the farthest range, ensuring reliable detection.   
For DCD, we observe slightly higher errors at close ranges (\(<2.5\) m) compared to mid-range distances (2.5–4.5 m). 
This is because within 2.5 m, the sensor captures only partial body heat signals, leading to missing information and minor degradation, though the error remains under 6.2 cm.
Notably, the lower error observed near 0.5 m is not an artifact of fewer samples, but reflects a geometric shift where ultra-close frames capture primarily upper-body regions rather than complex full-body reconstructions.
Beyond 2.5 m, DCD increases gradually as each pixel integrates radiation from larger body regions with mixed temperatures and emissivities, or non-human surfaces, complicating depth estimation.
Nevertheless, \sysname maintains high accuracy, with errors below 6.5 cm, well within human body scale.

\head{System cost and computational efficiency}
The \sysname prototype uses a Meridian MI0802M6S thermal array with an ESP32 MCU for data acquisition, resulting in a sensor-side hardware cost of about \$20 and enabling low-cost, scalable deployment.
mmWave radar boards have similar prices (\eg, TI IWR1443 and TI IWR6843, around \$20) but provide much lower spatial imaging resolution.
By contrast, many prior thermal sensing systems rely on high-resolution thermal cameras such as the FLIR A65C (\(>\$5{,}000\)) \cite{shinDeepDepthEstimation2023, shinDeepDepthEstimation2025a} and FLIR Boson 640 (\(>\$4{,}000\)) \cite{dingThermoHandsBenchmark3D2025}, as well as LiDAR units such as the Livox Mid-360 (\(\sim\$700\)), leading to substantially higher hardware cost.
In terms of computational performance, a five-minute continuous benchmark on a desktop PC (Intel i7-14700HX, RTX 4070) shows that \sysname runs at 15.02\,Hz (66.5\,ms per frame), exceeding the thermal array's 8\,Hz acquisition rate to confirm real-time capability, with 354\,MiB GPU memory usage and 7.41\,GFLOPs per inference.
This efficiency enables deployment on edge AI platforms like the Jetson Nano (\(\sim\$249\)), bringing the total cost of a fully autonomous node to approximately \$270.

\subsection{Micro-benchmark}
To further evaluate the robustness of \sysname, we conduct a series of micro-benchmark experiments.

\head{Cross-environment}  
To evaluate the cross-environment generalizability of \sysname, we test it on 13,459 samples collected from fully held-out environments (Env. 5–7).
As shown in \fig\ref{fig:cross_env}, \sysname achieves a mean F1 score of 0.985 and DCD of 2.30 cm in seen environments, and 0.967 F1 with 3.34 cm DCD in unseen environments. 
Although performance decreases slightly in unseen spaces, accuracy remains high, confirming robustness across different indoor layouts.

\head{Cross-user}  
\fig\ref{fig:cross_user} shows the results across seen and unseen users, where all samples from the unseen users are fully excluded from training and used only for testing (IDs 8–10).
For the 8 seen users, \sysname achieves an average F1 score of 0.991 and a mean DCD of 1.88 cm. For the 3 unseen users, performance remains comparable, with an average F1 score of 0.996 and DCD of 1.90 cm. 
This stable held-out-user performance suggests that \sysname learns transferable thermal-geometric representations rather than participant-specific patterns.

\head{Different user activities}  
\fig\ref{fig:cross_activity} reports performance under various activities, including walking, standing, sitting, lying, and cooking. 
F1 scores remain consistently high (0.96–1.00), while DCD varies between 1.4 cm (sitting) and 6.4 cm (walking), the latter due to motion blur during faster movements.
Overall, \sysname reliably adapts to different activities.

\head{Different clothing types}  
As shown in \fig\ref{fig:cross_clothes}, \sysname achieves robust results across seven clothing conditions (coat, jacket, T-shirt, mask, hat).
F1 scores are consistently above 0.97, with DCD ranging from 1.87 to 2.71 cm. 
This indicates strong resilience to clothing-related thermal variations.

\begin{figure*}[t]
\begin{minipage}{0.23\textwidth}
  \centering
  \includegraphics[width=0.99\textwidth]{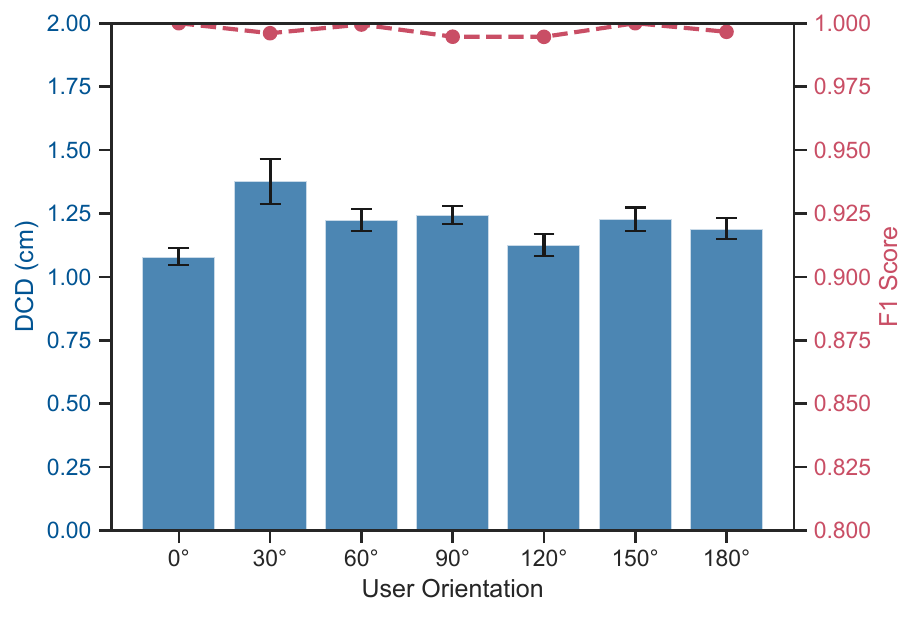}
  \vspace{-2\baselineskip}
  \caption{User orientation robustness.}
  \label{fig:cross_user_orientation}
\end{minipage}
\hfill
  \begin{minipage}{0.23\textwidth}
  \centering
  \includegraphics[width=0.99\textwidth]{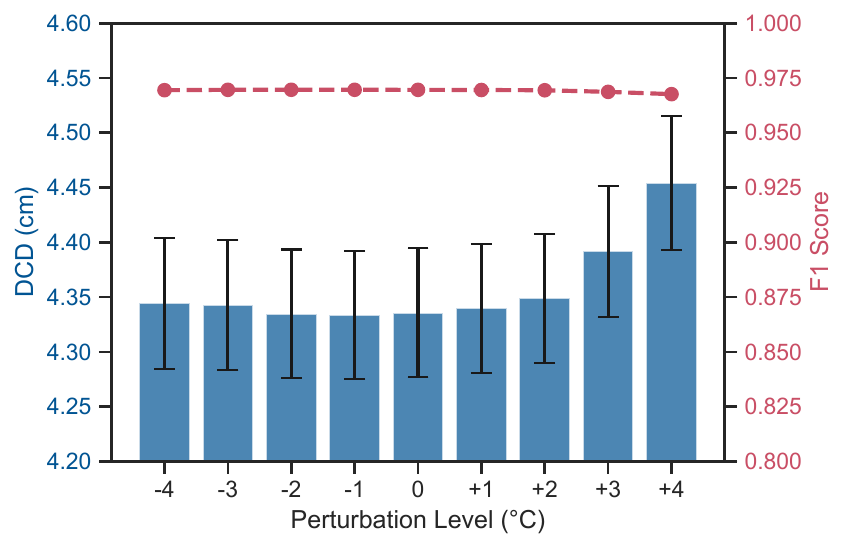}
  \vspace{-2\baselineskip}
  \caption{Temperature perturbation testing.}
  \label{fig:perturbation}
  \end{minipage}
\hfill
\begin{minipage}{0.23\textwidth}
  \centering
 \includegraphics[width=0.99\textwidth]{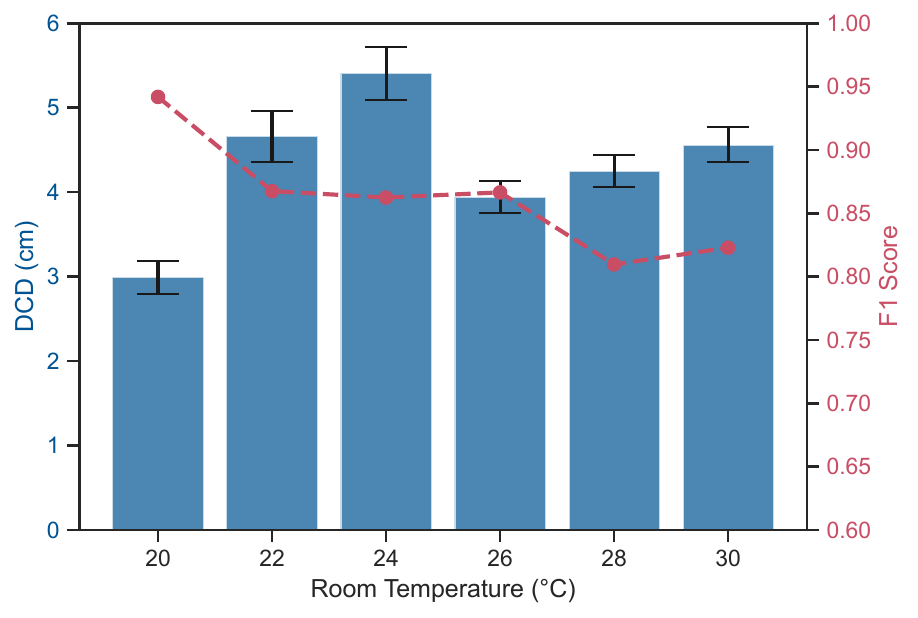}
  \vspace{-2\baselineskip}
  \caption{Room temperature robustness.}
  \label{fig:cross_room_temp}
\end{minipage}
\hfill
\begin{minipage}{0.23\textwidth}
  \centering
  \includegraphics[width=0.99\textwidth]{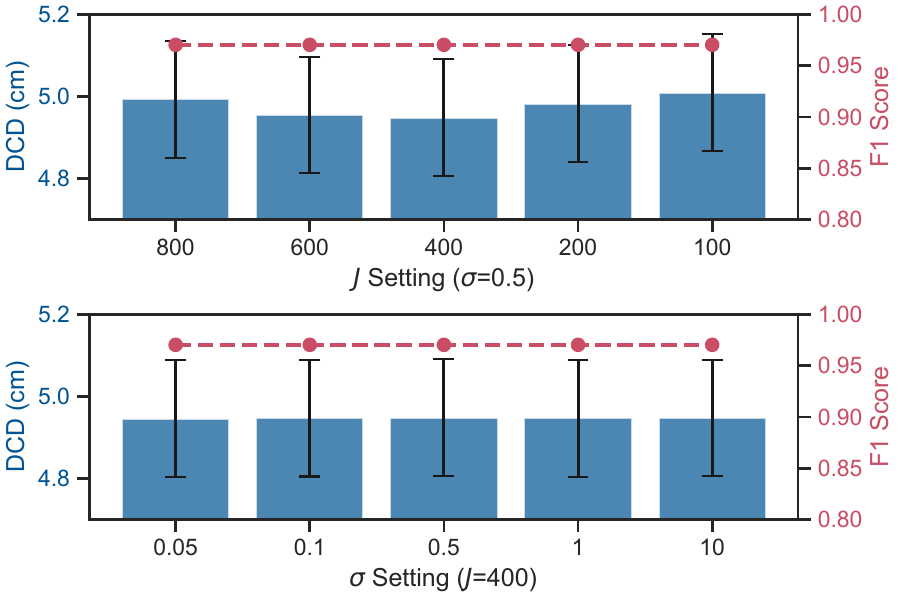}
  \vspace{-2\baselineskip}
  \caption{Impact of DIM parameters.}
  \label{fig:impact_DIM}
\end{minipage}
\end{figure*}

\begin{figure*}[t]
\begin{minipage}{0.23\textwidth}
  \centering
  \includegraphics[width=0.99\textwidth]{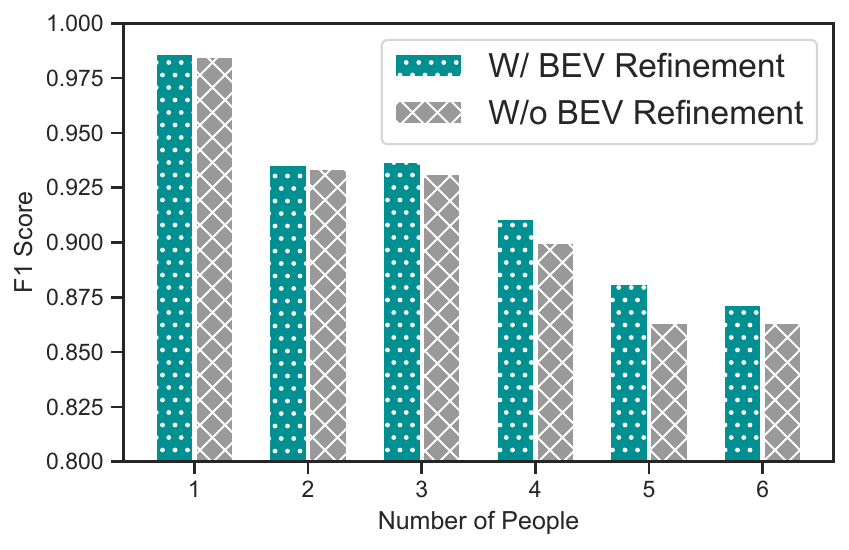}
  \vspace{-2\baselineskip}
  \caption{Effect of BEV refinement under varying numbers of users.}
  \label{fig:bev_ablation}
\end{minipage}
\hfill
\begin{minipage}{0.23\textwidth}
  \centering
  \includegraphics[width=0.99\textwidth]{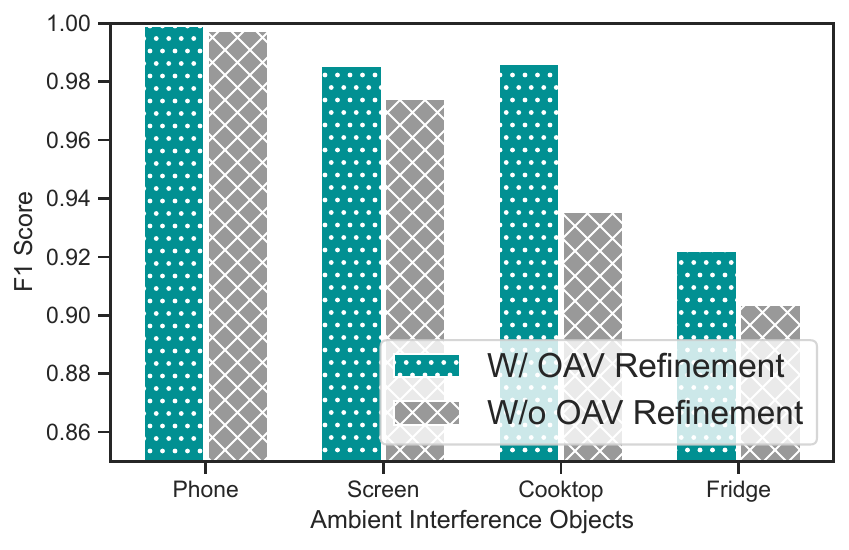}
  \vspace{-2\baselineskip}
  \caption{Effect of OAV refinement under different ambient thermal interferences.}
  \label{fig:aov_ablation}
\end{minipage}
\hfill
\begin{minipage}{0.23\textwidth}
  \centering
  \includegraphics[width=0.99\textwidth]{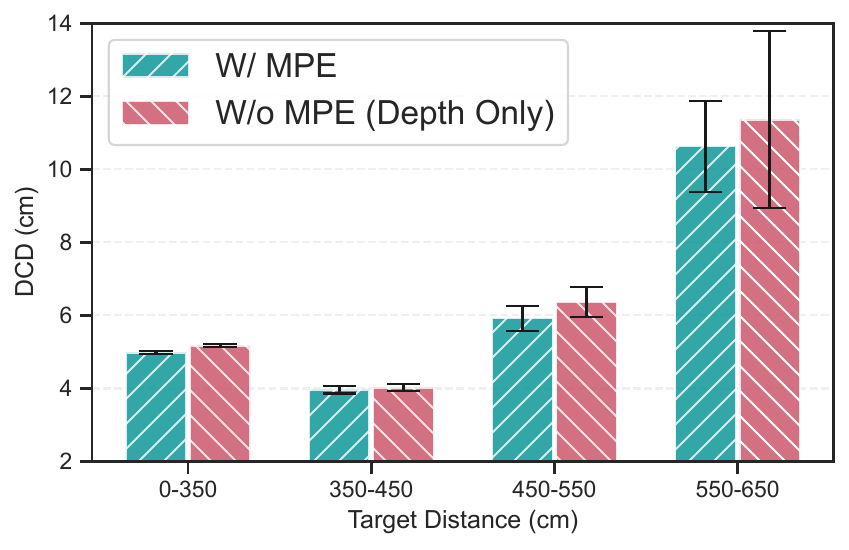}
  \vspace{-2\baselineskip}
  \caption{Effect of multi-primitive estimation (MPE) on point cloud reconstruction accuracy.}
  \label{fig:primitive_ablation}
  \end{minipage}
\hfill
\begin{minipage}{0.23\textwidth}
  \centering
  \includegraphics[width=0.99\textwidth]{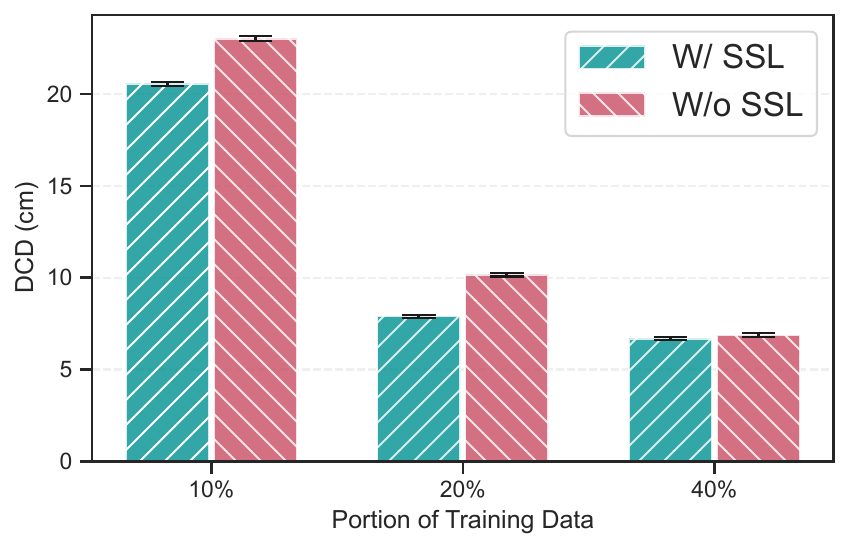}
  \vspace{-2\baselineskip}
  \caption{Effect of self-supervised pretraining with different amounts of labeled finetuning data.}
  \label{fig:ssl_ablation}
\end{minipage}
\end{figure*}

\head{Different occlusion levels}  
\fig\ref{fig:cross_occlusion} evaluates robustness under partial occlusion, where the lower body is blocked at different ratios using a 10~cm thermally opaque foam mattress as the occluder. Even with 80\% occlusion, \sysname maintains an F1 score of 0.989 and DCD of 2.23 cm, showing graceful degradation under incomplete body visibility.

\head{Different walking speeds}  
To evaluate the impact of motion blur, User~\#0 walked laterally at a distance of 2.5–3\,m across four speed levels (0.4, 0.7, 1.0, and 1.3\,m/s).
As shown in \fig\ref{fig:cross_walking_speed}, F1 scores remain near 1.0 and the DCD stays below 1.8\,cm across all speeds, confirming the system's robustness against motion blur.

\head{Different incident angles}  
\fig\ref{fig:cross_incident_angle} evaluates incident angles from 0° (frontal) to 45°. Performance remains stable up to 30° (F1 $>$ 0.99, DCD $<$ 2.0\,cm). At 45°, reduced visible area lowers performance (F1 = 0.926, DCD = 4.68\,cm), though accuracy remains practical for deployment.

\head{Different user orientations}  
\fig\ref{fig:cross_user_orientation} examines orientations at 0° (facing sensor) to 180° (back facing). F1 scores remain between 0.995 and 1.0, and DCD between 1.08 and 1.38 cm, regardless of orientation. These results demonstrate that \sysname remains highly accurate even when users are turned away from the sensor.

\head{Temperature perturbation robustness}
To assess robustness against model mismatch, we synthetically perturb test-time temperature maps with global offsets from $-4^\circ$C to $+4^\circ$.
As shown in \fig\ref{fig:perturbation}, \sysname remains stable under small perturbations ($\pm1$--$2^\circ$C), with negligible changes in DCD and F1 score, owing to its reliance on relative thermal contrasts and learned body structure priors. For larger perturbations ($\pm3$--$4^\circ$C), DCD increases more noticeably, as expected for temperature-sensitive depth estimation; however, the degradation remains limited, indicating graceful performance under substantial temperature mismatch.

\head{Room temperature impact}
We further evaluate \sysname under real ambient temperature variation using 16,187 samples collected at room temperatures from 20--30$^\circ$C.
As shown in \fig\ref{fig:cross_room_temp}, \sysname maintains stable reconstruction across this range, with an average DCD of 4.30\,cm and DCD between 2.99 and 5.41\,cm.
The average F1 score is 0.862, remaining above 0.86 from 20--26$^\circ$C and degrading modestly to 0.810--0.823 at 28--30$^\circ$C, indicating that \sysname generalizes reliably to common indoor room-temperature settings.

\head{DIM hyperparameter sensitivity}
We analyze the sensitivity of BEV refinement to the DIM hyperparameters $\sigma$ and $J$ by sweeping both around their default settings ($\sigma=0.5$, $J=400$).
As shown in \fig\ref{fig:impact_DIM}, varying $J$ from 100 to 800 and $\sigma$ from 0.05 to 10 changes DCD by less than 0.07\,cm (4.95--5.01\,cm), while the F1 score remains stable at 0.970 across all configurations.
These results indicate that \sysname's BEV refinement is robust to reasonable DIM design choices and that the selected defaults provide near-optimal performance.

\subsection{Ablation Study}
\label{subsec:ablation}

\head{Effect of BEV refinement}  
BEV refinement is designed to disentangle users in crowded scenes.
As shown in \fig\ref{fig:bev_ablation}, removing BEV refinement causes only a small F1 drop under sparse settings (1–3 users, –0.9\% on average), but a larger degradation under dense settings (4–6 users, –3.7\%).
This confirms that BEV refinement is particularly effective in resolving spatial overlap that cannot be reliably separated in the optical-axis view alone.

\begin{figure*}
\begin{minipage}{0.46\textwidth}
  \centering
 \includegraphics[width=0.88\textwidth]{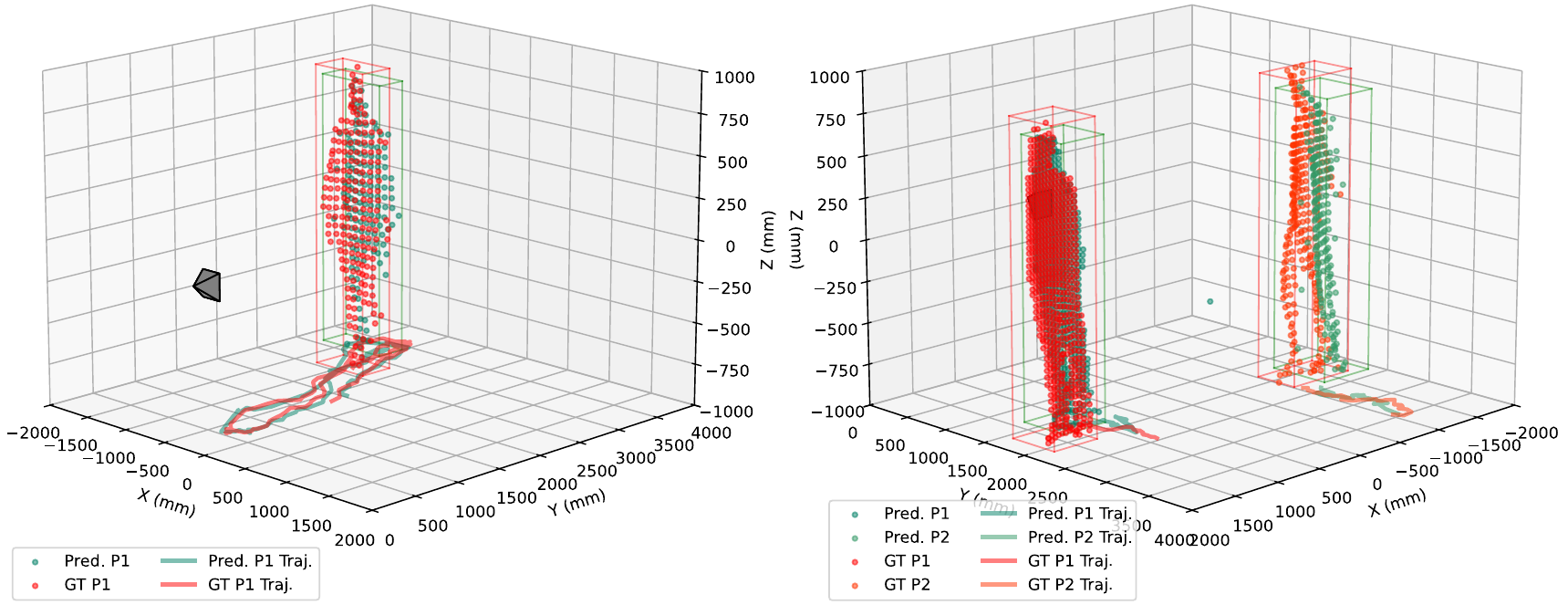}
  \vspace{-0.8\baselineskip}
  \caption{Indoor tracking. \rm{Examples of single- and multi-user trajectory estimation using \sysname outputs.}}
\label{fig:case_tracking}
\end{minipage}
\hfill
\begin{minipage}{0.48\textwidth}
  \centering
  \includegraphics[width=0.98\textwidth]{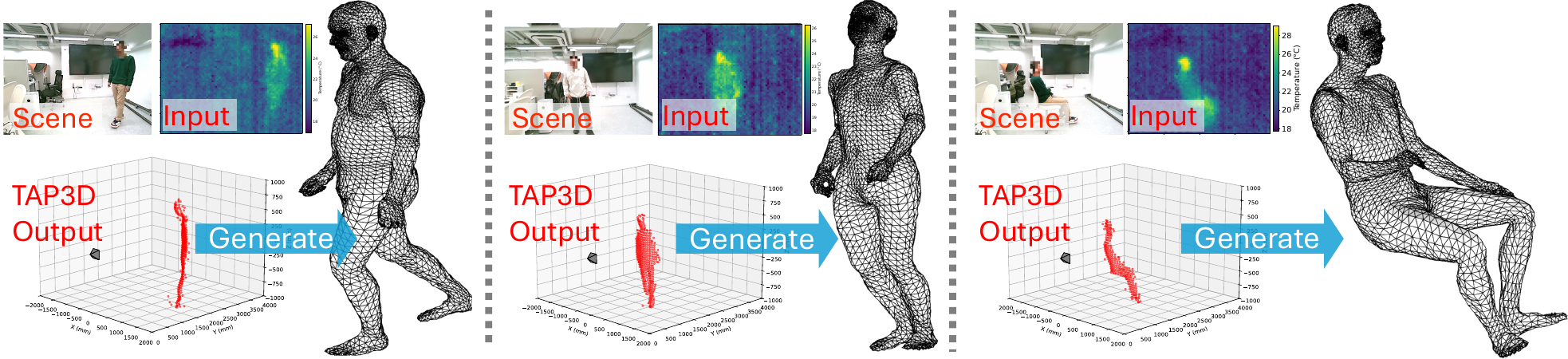}
 \vspace{-0.8\baselineskip}
  \caption{3D mesh recovery. \rm{Human meshes reconstructed from \sysname point clouds.}}
\label{fig:case_mesh}
\end{minipage}
\end{figure*}

\head{Effect of OAV refinement}  
OAV refinement suppresses ambient thermal interference.
As illustrated in \fig\ref{fig:aov_ablation}, incorporating OAV refinement improves detection performance across diverse interference sources.
Specifically, the F1 score increases from 0.974 to 0.986 for screen emissions (65-inch display), from 0.935 to 0.986 for cooking appliances (microwaves and gas stoves), and from 0.903 to 0.922 for refrigerators.
These gains validate that leveraging depth together with emissivity and surface temperature cues effectively reduces false positives caused by non-human heat sources.

\head{Effect of multi-primitive estimation}  
To evaluate the benefit of multi-primitive estimation, we compare \sysname with an ablated variant that estimates depth only and removes the forward physical model.
As shown in \fig\ref{fig:primitive_ablation}, multi-primitive estimation consistently yields lower DCD across all target distances.
Overall, this module reduces DCD by 5.59\%, demonstrating that jointly estimating depth with emissivity, surface temperature, and reflection effectively mitigates depth ambiguity inherent in thermal measurements.

\head{Effect of self-supervised training}
We isolate the contribution of physics-guided self-supervised learning (SSL) using a pretraining--finetuning protocol: 60\% unlabeled data for reconstruction-only pretraining and 40\% labeled data for finetuning.
Because \sysname requires explicit human detection, fully self-supervised end-to-end reconstruction is not yet supported.
As \fig\ref{fig:ssl_ablation} shows, SSL pretraining consistently improves accuracy across label budgets, reducing DCD by 11.93\% on average; fully self-supervised reconstruction remains future work.

\section{Case Study}
\label{sec:case}

\head{Fall detection}  
To demonstrate the potential of \sysname-generated point clouds for fall detection, we collect a dataset with three volunteers performing a total of 63 fall events. 
The thermal array sensor records temperature maps at 20 Hz, and each fall event is manually annotated.
To avoid test data leakage, six fall recordings are held out for testing. 
To enlarge the dataset, we deploy 12 sensors at different viewpoints for data collection.
The final dataset contains 756 fall samples and 5,973 non-fall samples.
For detection, we convert \sysname outputs into depth map sequences and train a ResNet-based binary classifier.
The system achieves 91.46\% accuracy, demonstrating the strong potential of \sysname point clouds for fall detection.

\head{Indoor human tracking}  
Indoor tracking provides location information for applications such as surveillance and elderly care \cite{zhang2010localization}. 
We evaluate TAP3D’s ability to track humans indoors using the following steps: (1) extract per-user point clouds and filter outliers with DBSCAN \cite{ester1996density}; (2) generate 3D bounding boxes per user; and (3) perform temporal association via maximum IoU matching across consecutive frames.  
As shown in \fig\ref{fig:case_tracking}, \sysname successfully tracks single- and multi-user trajectories, achieving a mean center distance error of 21.86\,cm across 144 one-minute recordings.
Compared with representative indoor tracking systems, this error is competitive with privacy-preserving RF solutions, including IR-UWB radar (17.7--23.3\,cm median) \cite{guoUWTrackingPassiveHuman2024}, single-target mmWave radar ($<$27\,cm) \cite{jiangIndoorHumanTracking2025}, and multi-subject mmWave radar (38--45\,cm) \cite{liuNovelHybridMachineLearning2026}, and is substantially lower than WiFi localization (32--83\,cm MAE) \cite{zhangRLocRobustIndoor2024a}.
Vision-based methods can reach 10.1\,cm with a monocular camera \cite{zhanMonocularPersonLocalization2025}, though under different sensing and deployment assumptions.
These comparisons suggest that \sysname point clouds demonstrate the potential for room-level indoor tracking in applications such as elderly care and smart buildings.

\head{3D mesh recovery}  
Human surface models provide richer shape information for behavior analysis. We train a ResNet-18 model \cite{he2016deep} to predict 3D human meshes from point clouds generated by \sysname. Ground-truth meshes are obtained from RGB images using the SMPL-X \cite{SMPL-X:2019}.
The dataset consists of 787 samples from three users performing walking and sitting, with 631 samples for training and the remainder for testing.  
As shown in \fig\ref{fig:case_mesh}, \sysname point clouds are effectively converted to detailed 3D meshes. 
The reconstructed models achieve a minimum matching distance of 4.87 cm, demonstrating high accuracy for human mesh recovery.
Although this study is exploratory and the limited dataset may introduce overfitting, the results highlight the strong potential of \sysname for downstream 3D mesh recovery and motivate future investigation.

\section{Limitations and Future Work}
\label{sec:limitation} 
While \sysname performs well indoors, important limitations remain.
Commodity thermal arrays have lower resolution than RGB-D cameras and are less robust outdoors under solar heating or weak human--background contrast. Currently, \sysname may degrade or fail in cases like severe multi-user overlap, varying heating sources, outdoor scenes, etc (Details in Appendix~\ref{app:failure_cases}).

\head{Beyond human body point cloud} 
\sysname focuses on reconstructing human point clouds by exploiting body heat signatures, eliminating the additional detection step required in LiDAR and mmWave systems \cite{zhangRFSearchSearchingUnconscious2023, zhangNeighborhoodenhanced3DHuman2024}.  
Future work could extend the general physical model in \S\ref{sec:physics} to non-human targets, enabling panoptic point cloud reconstruction.

\head{Point cloud density and resolution}  
\sysname reconstructs one 3D point per thermal pixel, resulting in hundreds of points per person given the resolution of current thermal arrays. This density exceeds compact radar and low-cost thermal ranging outputs but remains lower than RGB-D point clouds.
While this density is sufficient for coarse-to-medium geometric reconstruction, it is inherently limited by sensor resolution. Point density can be increased via interpolation or learning-based super-resolution techniques, which remains an interesting direction for future work.

\head{Precise multi-primitive estimation}
\sysname focuses on depth as the primary geometric primitive for 3D reconstruction.
The multi-primitive learning framework injects physical constraints to better isolate depth-related information, while precise multi-primitive estimation remains an interesting and challenging direction for future work.

\head{Multi-sensor integration}
Extending \sysname to multi-sensor setups could improve reconstruction accuracy in multi-user scenes.  
One approach is to align reconstructed point clouds in a shared coordinate frame using known sensor poses.

\head{Multi-modality fusion}  
Finally, \sysname offers a natural bridge for combining thermal arrays with complementary modalities such as LiDAR, radar, and depth cameras.  
Exploring advanced point cloud alignment methods will support richer, more robust multi-modal embodied AI applications.

\section{Related Works}
\label{sec:related_works}
We review literature on human body point cloud generation and thermal array–based human sensing.

\subsection{Human Body Point Cloud Generation}
Human body point clouds provide rich pose and shape information, making them a compelling choice for human sensing.
Existing methods for generating human body point clouds can be categorized by sensing modality: 

\head{Depth cameras} offer dense point clouds through direct depth measurements and are widely adopted in human-centric tasks \cite{sultaniKinect3DPoint2015, sunThreeDimensionalPointCloud2019}. 
They enable real-time applications such as gesture recognition \cite{leonVideoHandGestures2022}, activity recognition \cite{parkDepthCamerabasedHuman2016}, and surface geometry capture \cite{yuBodyFusionRealTimeCapture2017}. 
However, they are sensitive to illumination changes and raise privacy concerns; for example, Mozart \cite{xieMozartMobileToF2023} demonstrates texture recovery from ToF cameras, highlighting their privacy risks.

\head{LiDAR} provides long-range, high-fidelity 3D sensing under diverse lighting and environmental conditions.
LiDARCap achieves accurate 3D motion capture up to 30 meters via a hybrid kinematic-optimization framework \cite{liLiDARCapLongrangeMarkerless2022a}. 
HSC4D fuses LiDAR and IMU signals for human-centric scene reconstruction across indoor and outdoor domains \cite{daiHSC4DHumanCentered4D2022}. 
Sparse LiDAR measurements are refined to full-body meshes via graph transformers \cite{fanLiDARHMR3DHuman2023}, and LiveHPS introduces a spatiotemporal model to enhance robustness under occlusion \cite{renLiveHPSLiDARBasedSceneLevel2024}. 
LiDAR-Net further supports model generalization by providing annotated scans across large-scale indoor environments \cite{guoLiDARNetRealscanned3D2024}. 
However, LiDAR remains costly and constrained by eye-safety concerns \cite{hadlerRandomTestingReveals2013, zuclichOcularEffectsPenetrating1995}.

\head{Compact mmWave radar} offers an alternative that is lighting-invariant, and privacy-friendly. 
Yet, the generated point clouds are often sparse and noisy \cite{luSeeSmokeRobust2020}. 
To enhance fidelity, prior work applies either \textit{synthetic aperture radar} (SAR) or \textit{learning-based} densification techniques. 
MILLIPOINT performs coherent SAR imaging using self-tracked vehicle radars \cite{qian3DPointCloud2020a}, while handheld SAR systems address phase distortion without mechanical stabilizers \cite{liHighresolutionHandheldMillimeterwave2024}. 
On the learning side, mmPoint adopts a deformable encoder-decoder pipeline for dense reconstruction from single frames \cite{xie2023mmPoint}; RadarHD reconstructs LiDAR-like clouds from raw I/Q radar data \cite{prabhakaraHighResolutionPoint2023}; and mmMesh aligns sparse radar inputs with parametric body models for real-time mesh recovery \cite{xueMmMesh3DRealtime2021}. 
mmDiffusion advances this direction by leveraging temporal context in sequential signals, framing the problem as point cloud denoising through conditional diffusion \cite{xieMmDiffusionMmWaveDiffusion2025}. 
Despite these advances, mmWave radar still requires substantial post-processing to overcome intrinsic sparsity and suffers from cross-device interference \cite{brinon-arranzMethodologyAnalyzingImpact2021}.

In this regard, \sysname delivers a fully passive, thermal-assisted system for human body point cloud reconstruction, combining cost efficiency, high density, human sensitivity, and privacy preservation.

\subsection{Thermal Array-based Human Sensing}

Existing research on thermal array-based human sensing can be broadly categorized into the following two streams:  

\head{Task-oriented}  
Task-oriented methods process thermal array readings directly to accomplish application-specific objectives such as fall detection \cite{zhongMultiOccupancyFallDetection2021, rezaeiUnobtrusiveFallDetection2021, tatenoPrivacyPreservedFallDetection2020}, occupancy estimation \cite{chiduralaDetectionMovingObjects2022, chiduralaOccupancyEstimationUsing2021}, daily activity recognition \cite{muthukumarInfraredArraySensorBased2022, rezaeiUnobtrusiveHumanActivity2022, yinDeviceFreeHumanActivity2021, pollaActionRecognitionLowResolution2019}, human monitoring \cite{naserMultipleThermalSensor2022, perraMonitoringIndoorPeople2021}, gesture recognition \cite{tatenoHandGestureRecognition2019, vandersteegenLowLatencyHandGesture2020}, and indoor localization \cite{bouaziziLowResolutionInfraredArray2022, faulknerDeviceFreeLocalizationUsing2021, hevesiMonitoringHouseholdActivities2014}.  
These approaches typically treat the temperature map as a low-resolution gray-scale image for specific applications, without explicitly recovering intermediate spatial representations.

\head{Representation-oriented}  
Representation-oriented methods aim to extract fine-grained sensing representations from thermal array data before applying them to end tasks.  
For example, a U-Net-based approach \cite{naserAdaptiveThermalSensor2021} segments human silhouettes from wall-mounted thermal arrays, producing 2D body masks for applications such as human detection and activity recognition. However, this representation lacks range information.  
To incorporate distance cues, a subsequent method \cite{naserHumanDistanceEstimation2021} estimates human range by analyzing pixels in the bottom lines of the temperature map, which works only for single, front-facing users.  
Later, Zhang \etal \cite{zhangTADARThermalArraybased2024} enabled multi-user range estimation, generating a single-depth human body mask from the temperature map, with demonstrated applications in fall detection, occupancy monitoring, and sleep posture monitoring.  
Beyond whole-body representations, some studies target fine-grained hand pose recovery.  
FingerTrak \cite{huFingerTrakContinuous3D2020} and TAPOR \cite{Zhangtapor2025} target 3D hand tracking with thermal arrays; although physics-inspired, TAPOR still relies mainly on standard learning operators. \sysname scales to full-body 3D point cloud reconstruction via an explicit differentiable physical model.

To the best of our knowledge, \sysname is the first system to reconstruct dense 3D human point clouds from thermal array measurements.  
Unlike prior representations such as 2D masks \cite{naserAdaptiveThermalSensor2021} or single-depth estimates \cite{zhangTADARThermalArraybased2024}, our 3D point clouds provide substantially richer geometric detail.  
This enables downstream tasks previously infeasible, such as full-body mesh reconstruction, and can be readily leveraged by existing point cloud models.  
Moreover, \sysname facilitates integration of thermal arrays with other modalities, such as radar and LiDAR, using point clouds as a bridge representation for cross-modality alignment.

\section{Conclusion}
\label{sec:conclusion}

We presented \sysname, the first system to transform body heat signatures into dense 3D human point clouds using a single low-cost thermal array sensor.  
With a physics–informed design, \sysname disentangles depth, emissivity, temperature, and reflection through a multi-primitive estimation module, and achieves robust interference suppression and multi-user separation via a geometric perspective fusion module.  
Extensive experiments on a large-scale dataset of over 160K samples demonstrate high geometric accuracy and reliable detection across diverse conditions.  
Case studies on fall detection, indoor tracking, and 3D mesh recovery further showcase its potential as a versatile foundation for privacy-preserving, fully passive human sensing, enabling rich human-centric applications.

\begin{acks}
This work is supported by the Hong Kong RGC GRF under grant No. 17211725 and HKU Seed Fund for Collaborative Research under grant No. 2507263051. 
\end{acks}

%% file: appendix.tex
\appendix
\section{Detailed Derivation of the \sysname Physics Model}
\label{app:physics_model}

This appendix provides the full derivation of the physics model underlying \sysname, extending the concise formulation in \S\ref{sec:physics}. We follow classical thermography theory~\cite{mollmannInfraredThermalImaging2018} to relate human-body primitives to thermal array measurements.

\subsection{Forward Physical Model}

We first construct a forward model that describes how the position and thermal properties of a target region affect the temperature reported by a thermal array pixel.  
As illustrated in \fig\ref{fig:theory_diagram}, each pixel $A_p$ on the image plane receives thermal radiation from a corresponding surface region $A_t$ on the human body and outputs an estimated temperature value. Aggregating all pixels yields the full temperature map.

To distinguish different radiation sources, we use subscripts $t$, $i$, and $p$ for quantities associated with the target region $A_t$, ambient interference regions $A_i^{\cdot}$, and the sensor pixel $A_p$ (receiver), respectively.

\head{Atmospheric transmittance}
To determine the three dimensional location of $A_t$, the key quantity is its depth (distance along the $X$-axis) relative to the sensor. The image-plane coordinates $(Y,Z)$ are determined by the sensor FoV and spatial resolution, while depth must be inferred from the physical interaction between emitted radiation and the atmosphere.  
Thermal radiation emitted from the target and transmitted through air experiences attenuation governed by the Bouguer–Lambert–Beer law, which yields an exponential decay of transmittance with distance~\cite{nagaseShapeThermalRadiation2022}.  
Although major air components (N\textsubscript{2}, O\textsubscript{2}) are mostly transparent in the LWIR band, heteronuclear molecules such as H\textsubscript{2}O and CO\textsubscript{2} exhibit strong absorption~\cite{mollmannInfraredThermalImaging2018}.  
The spectral atmospheric transmittance over distance $r$ is modeled as
\begin{equation}
    \tau_{\lambda}(r) = \exp(-\gamma_{\lambda} r),
\end{equation}
where $\gamma_{\lambda}$ is the spectral attenuation coefficient and $\lambda$ is the wavelength.

\head{Spectral exitance of the target region}
Given the location $\overrightarrow{P}_t \in \mathbb{R}^3$ and surface normal $\vec{a}_t$ of the target region $A_t$, the original spectral exitance $M_{\lambda,t}$ consists of (i) self-emission from $A_t$ and (ii) reflected radiation from ambient interference regions $A_i^k$ (e.g., other body parts or nearby devices):
\begin{equation}
M_{\lambda, t} = \epsilon_{\lambda, t} M_{\lambda}^{\text{BB}}(T_t)
+ (1 - \epsilon_{\lambda, t}) \sum_{k=1}^{K} \tau_{\lambda}(r_{k \to t}) V_{k \to t} M_{\lambda, k}.
\end{equation}
The first term models self-emission governed by surface temperature $T_t$ and spectral emissivity $\epsilon_{\lambda,t}$; $M_{\lambda}^{\text{BB}}(\cdot)$ is the Planck spectral radiance function~\cite{mollmannInfraredThermalImaging2018}.  
The second term accounts for reflected radiation from $K$ ambient regions at locations $\overrightarrow{P}_i^k$, $k=1,\dots,K$. Here,
$r_{k \to t} = \|\overrightarrow{P}_t - \overrightarrow{P}_i^k\|$ is the distance between $A_k$ and $A_t$, $\tau_{\lambda}(r_{k \to t})$ is given above, and
$V_{k \to t}$ is the view factor describing geometric visibility~\cite{baoHeatassistedDetectionRanging2023}, approximated as $V_{k \to t} = \xi_t A_t$ with tunable $\xi_t \in [0,1]$.  
Finally, $M_{\lambda,k}$ is the spectral exitance of the $k$-th interference region.

We adopt three simplifying assumptions that are practical for 3D human point cloud generation and validated experimentally:
\begin{enumerate}
    \item The target region $A_t$ behaves as a Lambertian radiator with direction-independent emissivity $\epsilon_{\lambda,t}$, yielding $L_{\lambda,t} = M_{\lambda,t}/\pi$.
    \item Ambient interference regions are treated as uniform surfaces contributing identical differential view factors to $A_t$.
    \item Radiation transfer from $A_t$ to interference regions and inter-reflections among interference regions are neglected, excluding higher-order effects.
\end{enumerate}

\head{Optical transfer}
By combining the pinhole camera model with atmospheric attenuation, the spectral radiation flux $\Phi_{\lambda,p}$ arriving at pixel $A_p$ is
\begin{equation}
    \Phi_{\lambda,p}
    = G_{\lambda} \, \tau_{\lambda}(\|\vec{P}_t\|) \, L_{\lambda,t}
      \frac{\pi D^2 \cos^4 \alpha}{4 f^2} \, A_p,
\end{equation}
where $G_{\lambda}$ is the optical gain, $\|\vec{P}_t\|$ is the distance from $A_t$ to the sensor, $\tau_{\lambda}(\|\vec{P}_t\|)$ is as above, $L_{\lambda,t}$ is the spectral radiance of $A_t$, $\alpha$ is the angle between the optical axis and the projection line, and $f$ and $D$ are the lens focal length and aperture diameter.

\head{Thermoelectric conversion and temperature readout}
Each thermal array pixel employs a thermopile that converts total received spectral radiation
$\Phi_p = \int_{\lambda_1}^{\lambda_2} \Phi_{\lambda,p} \, d\lambda$
over the sensor’s spectral response band $(\lambda_1,\lambda_2)$ into a voltage signal $V_p$ according to the thermoelectric response model~\cite{budzierThermalInfraredSensors2011}:
\begin{equation}
V_p = N \kappa \rho \frac{\Phi_p}{G_{\text{th}}},
\end{equation}
where $N$ is the number of thermocouples, $\kappa$ is the Seebeck coefficient~\cite{mollmannInfraredThermalImaging2018}, $\rho$ is the absorbance, and $G_{\text{th}}$ is the thermal conductance.  
Commercial thermal arrays then convert $V_p$ to an estimated temperature $\hat{T}_p$ via the Sakuma–Hattori equation~\cite{Sakuma1983Establishing,sheininProjectingTrackableThermal2024}, with sensor-specific calibration coefficients.

\subsection{Inverse Problem Formulation}

The forward physical model describes how sensor parameters, atmospheric properties, the target user, and ambient interferences jointly determine the thermal array output. For 3D human point cloud estimation, we simplify this model and formulate the inverse problem accordingly.

We adopt two standard assumptions:  
(1) Consistent with~\cite{zhangTADARThermalArraybased2024,baoHeatassistedDetectionRanging2023}, all surface regions, including $A_t$ and $A_i^k$, behave as Lambertian gray bodies with wavelength-independent emissivity $\epsilon$.  
(2) For wavelength-dependent quantities such as $G_{\lambda}$ and $\gamma_{\lambda}$, we use band-averaged values $G$ and $\gamma$ over the sensor’s operational spectrum.

We then categorize variables as:  
(1) unknown target primitives: $\overrightarrow{P}_t$, $T_t$, $\epsilon_t$;  
(2) fixed parameters: $\overrightarrow{P}_i^k$, $M_{\lambda,k}$, $\xi_t$, $G$, $D$, $f$, $A_p$, $\zeta = N \kappa \rho / G_{\text{th}}$, and $\gamma$;  
(3) observed quantities: $\hat{T}_p$.

The main text (\S\ref{sec:physics}) presents the resulting inverse model:
Eqs.~\eqref{eq:physic_final}–\eqref{eq:physic_v2} compactly define the physics-constrained inverse problem that \sysname solves to recover dense human point clouds from thermal array measurements.

\section{Training Objective}
\label{app:training}

This appendix provides the full training objective used in \S\ref{subsec:training}.
The dataset provides thermal array readings together with depth maps and RGB images from a co-located stereo camera, which are processed to obtain the ground-truth depth map $D_g$ and human indication map $M_g$, where $\tilde{M}_g = \mathbf{1}_{\{M_g > 0\}}$ denotes the binary foreground mask.

The \textit{multi-primitive estimation module} is trained with a reconstruction loss for self-supervision,
\[
L_{\text{recon}}(\hat{T}, T_{\text{recon}}; \tilde{M}_g) = \text{Huber}(\tilde{M}_g \odot \hat{T}, \ \tilde{M}_g \odot T_{\text{recon}}),
\]
a depth loss supervising the predicted depth map $D_t$:
\[
L_{\text{depth}}(D_t, D_g; \tilde{M}_g) = \text{Huber}(D_t, \ \tilde{M}_g \odot D_g),
\]
and an indication map loss:
$L_{\text{ind}}(M_t, M_g) = \text{Huber}(M_t, M_g)$.

The \textit{geometric perspective fusion module} is trained with two losses:
$L_{\text{OAV}}(M_{\text{OAV}}, M_g) = \text{BCE}(M_{\text{OAV}}, M_g)$ for OAV segmentation, where BCE is the binary cross-entropy loss; and $L_{\text{BEV}}(O_{\text{BEV}}, D_g, M_g) = \text{BCE}\!\left(O_{\text{BEV}}, \ \text{DIM}(M_g \odot D_g)\right)$
for BEV refinement, where $\text{DIM}(\cdot)$ is the Differentiable Index Mapping in \S\ref{subsec:multi-view}.

The overall objective is
\[
\mathcal{L} = \lambda_1 L_{\text{recon}} + \lambda_2 L_{\text{depth}} + \lambda_3 L_{\text{ind}} + \lambda_4 L_{\text{OAV}} + \lambda_5 L_{\text{BEV}},
\]
with weighting factors $\lambda_1, \lambda_2, \lambda_3, \lambda_4, \lambda_5$ balancing the terms.

\begin{figure*}[t]
    \includegraphics[width=0.86\textwidth]{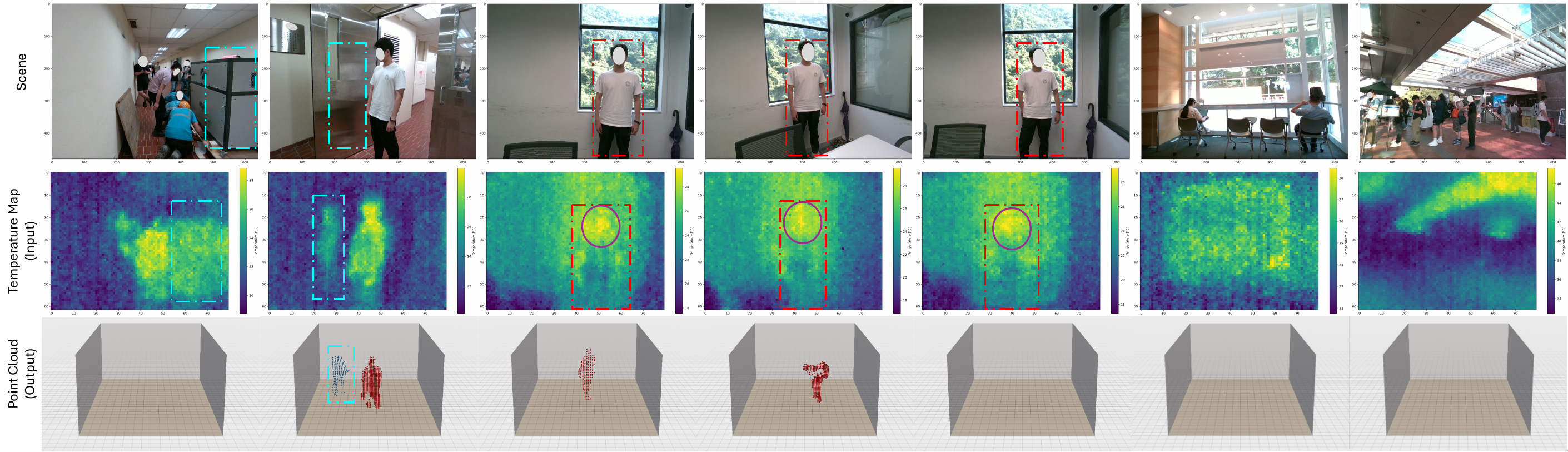}
    \vspace{-0.8\baselineskip}
    \caption{Representative failure cases. \rm{Each column shows the RGB scene (1st row), thermal input (2nd row), and predicted point cloud (3rd row). Columns illustrate (1) multi-user overlap with nearby interference (cyan dashed box), (2) false positives from reflective surfaces (metal door, cyan dashed box), (3--6) unstable or failed detection before sun-heated windows, and (7) outdoor failure under strong ambient thermal radiation.}}
    \label{fig:failure_case}
\end{figure*}

\section{Dataset, Baselines, and Metrics}
\label{app:dataset_baselines}

This appendix provides the full dataset, baseline, and metric details used in \S\ref{subsec:dataset_baselines}.

\head{Dataset}
We conduct experiments on a large-scale dataset collected in eight indoor environments (see \fig\ref{fig:setup_vis}) from 11 volunteers (four female and seven male), aged 19--28 years, with heights ranging from 165 cm to 187 cm.
All procedures were approved by our institution’s IRB.
The dataset contains over 160{,}000 samples, each with a synchronized temperature map, 3D point cloud ground truth, and human mask.
Ground-truth point clouds are obtained using a co-located RealSense D455 depth camera, while human masks are generated with Detectron2 \cite{wu2019detectron2}.
To accelerate data collection, samples are recorded at 8 Hz in one-minute segments, yielding 328 segments in total.
To prevent cross-segment data leakage, we split the dataset at the segment level: the training set contains 88{,}631 samples (with 20\% reserved for validation), and the test set contains 80{,}011 samples.

\head{Baselines}
To the best of our knowledge, \sysname is the first thermal array human point cloud reconstruction system, so no direct baselines exist.
We construct representative baselines by adapting SOTA methods from two domains for thermal array inputs: (i) single-view RGB--to--point cloud reconstruction, and (ii) thermal image--based depth estimation followed by back-projection.
Rather than using off-the-shelf weights, we train all baseline models from scratch on the complete \sysname training split.

\begin{itemize}[noitemsep, topsep=0pt, leftmargin=*]
\item \textbf{RGB2Point}~\cite{lee2025rgb2point} is a SOTA single-view image--to--point cloud reconstruction model based on Transformers.
To adapt it to thermal arrays, we upsample each thermal frame to $224\times224$, replicate it to three channels, and feed it to the original RGB2Point architecture.

\item \textbf{NeWCRF}~\cite{yuan2022neural} is a SOTA monocular depth estimation model that combines a ViT encoder with a CRF-based decoder and has shown strong results on depth benchmarks~\cite{shinDeepDepthEstimation2023, shinDeepDepthEstimation2025a}.
We apply NeWCRF to interpolated thermal frames ($480\times640$) to predict depth maps, back-project them into 3D point clouds, and then resize the outputs to the original thermal resolution ($62\times80$).
\end{itemize}

\head{Metrics}
We adopt two metrics:
(1) \textit{Directed Chamfer Distance (DCD).} DCD measures geometric accuracy as the average shortest distance from predicted points $P$ to ground truth $Q$:
$d_{\text{dir}}(P \to Q) = \frac{1}{|P|} \sum_{p \in P} \min_{q \in Q} \| p - q \|_2.$
In practice, we compute the symmetric Chamfer distance $d_{\text{CD}}(P, Q)$ using PyTorch3D \cite{ravi2020pytorch3d}, then report DCD as
$
\text{DCD} = \sqrt{\tfrac{1}{2} d_{\text{CD}}} \quad [\text{cm}] .
$
Unlike the symmetric version, DCD is more intuitive since it directly reflects prediction-to-ground-truth error.
(2) \textit{F1 Score.} For user detection, we compute the F1 score by evaluating precision and recall over bounding boxes extracted from the predicted and ground-truth human indication maps at an IoU threshold of 0.5.

\section{Representative Failure Cases}
\label{app:failure_cases}

\fig\ref{fig:failure_case} illustrates representative failure modes that clarify the operating conditions of \sysname.
Each column shows the RGB scene (top), the input thermal map (middle), and the predicted point cloud (bottom).

\head{Severe multi-user overlap}
When multiple users overlap in the scene (Col.~1), their OAV signatures can remain connected even after BEV refinement. This issue is further exacerbated when nearby heat sources introduce additional interference (cyan dashed boxes), ultimately leading to missed detections.

\head{Spatially varying heat sources}
Highly reflective surfaces and sun-heated structures create localized thermal fields that violate the global interference assumptions used during training.
When a user stands near a metal door, reflected body heat can produce false-positive detections (Col.~2).
When a user stands before a sun-heated window, human--background contrast becomes unstable across frames: although the head region often appears warmer than the window (purple boxes in the temperature maps), the lower body may appear cooler, yielding inconsistent signatures.
Consequently, reconstruction may succeed (Col.~3), degrade (Col.~4), or fail entirely (Cols.~5--6) as window radiation increasingly overwhelms the body signature.

\head{Interference-heavy and outdoor scenes}
In outdoor settings on sunny days (Col.~7), thermal radiation from ambient objects and the surrounding air can dominate the temperature map, leaving insufficient human contrast for reliable detection.
Together, these cases indicate that \sysname is most reliable in indoor environments with moderate thermal interference.

\head{Potential solutions}
Many of these failures stem from the limited spatial resolution and temperature sensitivity of commodity thermal arrays.
Higher-resolution, higher-sensitivity sensors could improve contrast between users and between humans and the background, making overlapping people and weakly contrasting scenes easier to separate.
At the algorithm level, enforcing temporal consistency and human-body structural priors---for example, via mesh or kinematic constraints applied as post-processing---may further stabilize detection and reconstruction in the challenging scenarios above, including multi-user overlap and spatially varying heat sources.